\RequirePackage{fix-cm}
\documentclass[twocolumn]{svjour3}          
\smartqed  
\usepackage{graphicx}
\usepackage[utf8]{inputenc} 
\usepackage[T1]{fontenc}    
\usepackage[hidelinks]{hyperref}
\hypersetup{
  colorlinks   = true, 
  urlcolor     = blue, 
  linkcolor    = blue, 
  citecolor   = red 
}
\usepackage{url}            
\usepackage{booktabs}       
\usepackage{amsfonts}       
\usepackage{amsmath}
\usepackage{nicefrac}       
\usepackage{microtype}      
\usepackage{tabularx}
\usepackage{multirow}
\usepackage{appendix}
\usepackage{lipsum}
\usepackage{color,soul}
\usepackage[export]{adjustbox}
\usepackage{xspace}
\usepackage{caption, sidecap}
\usepackage{subcaption}
\usepackage[dvipsnames]{xcolor}

\newcommand{\changeS}[1]{\textcolor{black}{#1}}

\newcommand{\changeF}[1]{\textcolor{White}{#1}}

\newcommand{\model}{HD\textsuperscript{2}S\xspace}

\begin{document}

\title{Hierarchical Domain-Adapted Feature Learning for Video Saliency Prediction}



\author{G. Bellitto$^*$ \and F. Proietto Salanitri$^*$ \and S. Palazzo$^+$ \and F. Rundo \and D. Giordano \and C. Spampinato}


\institute{G. Bellitto, F. Proietto Salanitri, S. Palazzo, D. Giordano, C. Spampinato \at
              PeRCeiVe Lab - University of Catania \\
              Tel.: +39-095-7387905\\
              \url{www.perceivelab.com}           
             \emph{Present address:} of F. Author  
           \and
           F. Rundo \at
              STMicrolectronics, ADG Central R\&D, Catania, Italy\\
            $^*$ Contribute equally.
            $^+$ Corresponding author.
}

\date{}

\maketitle

\begin{abstract}
\changeS{In this work, we propose a 3D fully convolutional architecture for video saliency prediction that employs hierarchical supervision on intermediate maps (referred to as \emph{conspicuity maps}) generated using features extracted at different abstraction levels.
We provide the base hierarchical learning mechanism with two  techniques for \textit{domain adaptation} and \textit{domain-specific learning}. For the former, we encourage the model to \changeS{unsupervisedly} learn hierarchical general features using gradient reversal at multiple scales, to enhance generalization capabilities on datasets for which no annotations are provided during training.
As for domain specialization, we employ domain-specific operations (namely, priors, smoothing and batch normalization) by specializing the learned features on individual datasets in order to maximize performance. 
The results of our experiments 
show that the proposed model yields state-of-the-art accuracy on supervised saliency prediction.  When the base hierarchical model is empowered with domain-specific modules, performance improves, outperforming state-of-the-art models on three out of five metrics on the DHF1K benchmark and reaching the second-best results on the other two.
When, instead, we test it in an unsupervised domain adaptation setting, by enabling hierarchical gradient reversal layers, we obtain performance comparable to supervised state-of-the-art.\\
Source code, trained models and example outputs are publicly available at \url{https://github.com/perceivelab/hd2s}.}
\end{abstract}

\section{Introduction}
Video saliency detection is the task of \changeS{predicting} human gaze fixation when perceiving dynamic scenes, and it is typically carried out by estimating spatio-temporal saliency maps from an input video sequence. 
Saliency detection, in general, can be seen as the upstream processing step of multiple applications that include object detection~\cite{Girshick_2015_ICCV}, behavior understanding~\cite{lim2014crowd,lu2017crowd}, video surveillance~\cite{li2007fast,shao2005tracking,guraya2010predictive,yubing2011spatiotemporal} and video captioning~\cite{nguyen2013static,wang2018spotting,chen2018saliency}.
Existing video saliency detection methods generally apply single-image saliency estimation on individual frames, and combine the results with recurrent layers to temporally model frame-level features. However, the two separate analysis stages in these models make them unable to fully capture spatio-temporal features simultaneously. Recently, 3D fully-convolutional models have addressed this limitation by progressively aggregating spatio-temporal cues, achieving state-of-the-art performance on standard benchmarks. 
For example, TASED-Net~\cite{min2019tased} adopts a standard encoder-decoder architecture, as largely used in semantic segmentation tasks~\cite{ronneberger2015u,badrinarayanan2017segnet,noh2015learning}, that learns a compact spatio-temporal representation, and feeds it to a decoder subnetwork to perform saliency prediction. \changeS{While these methods perform well, saliency prediction is constrained by the aggregated representation learned at the model's bottleneck. This leads to learn representations that are more specific to the training data distribution, consequently limiting model generalization.}\\
\begin{figure}[ht]
    \centering
    \includegraphics[width=\columnwidth]{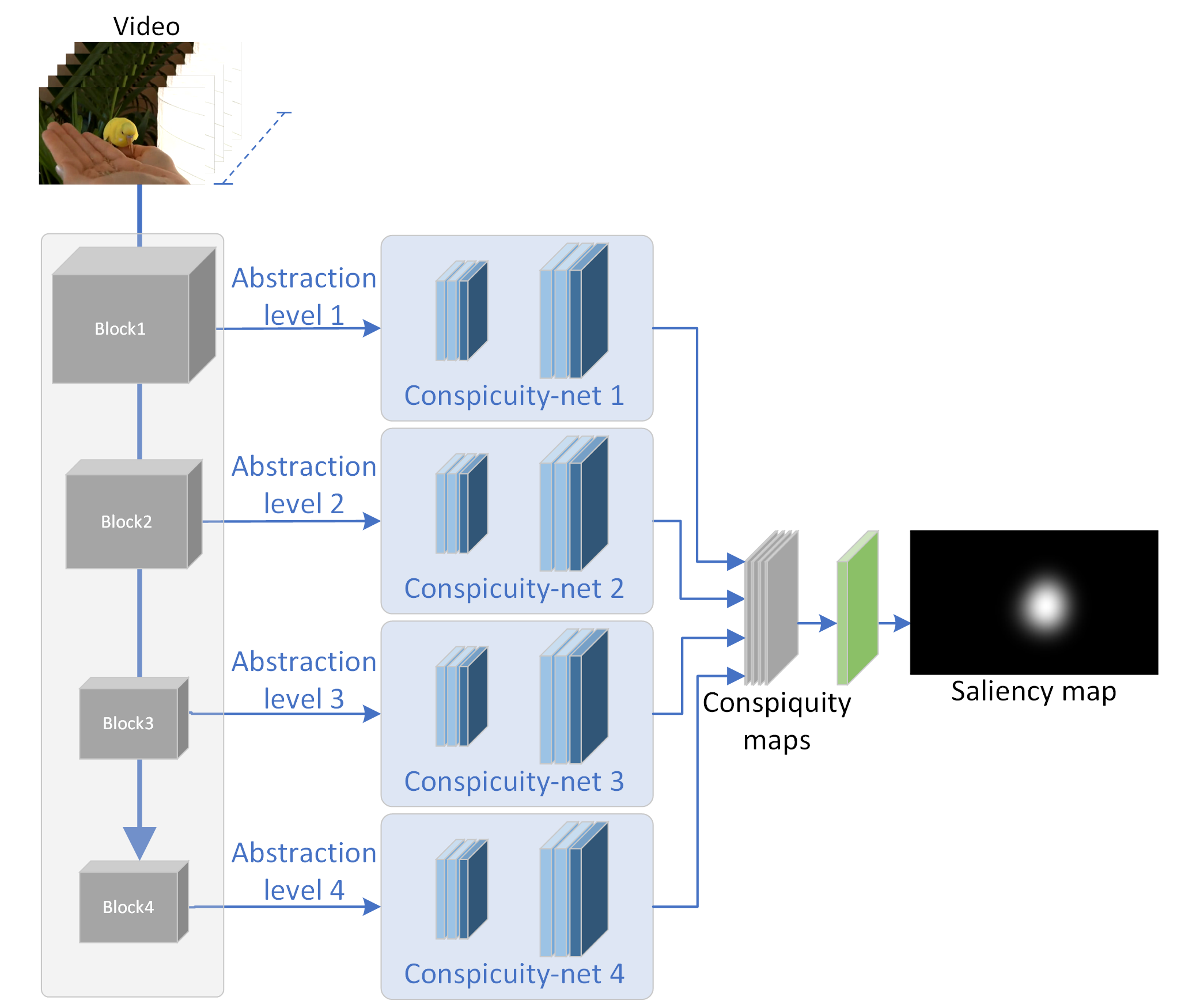}
    \caption{\changeS{\textbf{HD\textsuperscript{2}S overview}. Our proposed model generates multiple intermediate saliency maps by using features extracted at different abstraction levels, and combines them to predict the output map. We refer to the intermediate saliency maps as \textit{conspicuity maps}.}}
    \label{fig:overview}
\end{figure}

Following the success of 3D convolutional architectures, \changeS{in this paper we propose a model based on \emph{Hierarchical Decoding for Dynamic Saliency} prediction --- \emph{HD\textsuperscript{2}S} ---} that, instead of using a compact spatio-temporal representation as in~\cite{min2019tased}, generates multiple saliency maps by using features learned at different abstraction levels and then combines them to compute the final output. We refer to the intermediate saliency maps as \emph{conspicuity maps}, as the employed architecture recalls the multi-scale model proposed in~\cite{itti1998model}. 
Using representations extracted at different abstraction levels (from shallow to deeper) allows the model to learn both generic (and more dataset-independent) and dataset-specific features. The twofold advantage we obtain is to enhance performance on a specific dataset and, at the same time, to improve adaptation capabilities. \\
Our approach takes inspiration from DVA~\cite{wang2018deep}, but extends it to the video domain by learning spatio-temporal cues for predicting visual saliency. 
More specifically, HD$^2$S, shown in Fig.~\ref{fig:overview}, is a 3D fully-convolutional network that employs an ensemble of multiple prediction models, each producing a \emph{conspicuity-like} map at a specific abstraction level, for better saliency estimation.

As an additional contribution, we tackle the problem of generalization for video saliency \changeS{prediction}. Indeed, state-of-the-art methods lack domain adaptation capabilities and require a mandatory fine-tuning step to perform well on datasets that they were not trained on. As the deep learning community is moving to build more generalizable models, we argue that this holds, even more so, for saliency \changeS{prediction} research, given its fundamental nature in an artificial vision pipeline.
To address this issue, our saliency \changeS{prediction} network is provided with a multi-scale domain adaption mechanism, based on gradient reversal~\cite{ganin2016domain}, that encourages the model to learn \textit{domain-independent} features. In particular, each abstraction level of HD$^2$S is provided with a gradient reversal layer that prevents the learned representation from becoming dataset-specific. 

\changeS{We also address the opposite problem, i.e., \textit{domain-specific learning}, by adding to the model some dataset-specific modules whose parameters are learned in order to maximize performance on a given dataset.}\\
We carry out extensive experiments testing of HD$^2$S on multiple video saliency benchmarks (DHF1K \cite{wang2018revisiting}, UCF Sports \cite{marszalek2009actions,soomro2014action}, Hollywood2 \cite{6942210}) obtaining state-of-the-art performance and outperforming existing models. \changeS{Performance that are boosted, as expected, when domain-specific learning is enabled}. We also test thoroughly the domain adaptation capabilities of HD$^2$S to datasets for which no annotations are available during training. Our model shows remarkable results, achieving performance comparable to state-of-the-art models that, instead, are trained (or fine-tuned) on those datasets in a standard supervised fashion.

\section{Related work}
Saliency detection has been long investigated in AI and computer vision research. In general, saliency models can be categorized in: \textit{saliency prediction}~\cite{wang2019revisiting} approaches that attempt to predict the fixation points of a human observer \changeS{during free-viewing (e.g., they aim to predict where people look at in a scene)}, and \textit{salient object detection}~\cite{liu2010learning} methods that, instead, focus on assessing the saliency of pixels w.r.t. objects of interest \changeS{(e.g., they aim to separate the salient objects from the background).} 
Saliency methods can be further categorized according to whether they process still images (static saliency) or videos (dynamic saliency).

Static saliency has been studied for decades. Initial models, biologically-inspired~\cite{itti1998model} and employing hand-crafted features, were followed by recent CNN-based attempts~\cite{huang2015salicon,pan2016shallow,pan2017salgan,Kummerer_2017_ICCV,wang2018deep,fan2018emotional,8400593,8866748,kroner2020contextual,jia2020eml} that yield superior performance, rapidly becoming state of the art for static saliency \changeS{prediction}. S 
\changeS{To overcome the lack of large eye fixation datasets, CNN-based static methods rely mainly on image classification models, as backbone, exploiting their capability to extract features useful for other visual tasks. Different encoder-decoder architectures with various strategies to combine the extracted features have then been proposed. The release of larger dataset for saliency, such as MIT300~\cite{judd2012benchmark},  SALICON~\cite{jiang2015salicon}, and CAT2000~\cite{borji2015cat2000}, led to a performance gain.
DeepGaze II~\cite{Kummerer_2017_ICCV} investigated the benefit of employing low- and high-level features in saliency prediction. Similarly, ML-NET~\cite{cornia2016deep} proposed to combine low- and high-level features at the bottleneck, while~\cite{kroner2020contextual} concatenates the outputs from several layers and processes them with multiple convolutional layers with different dilation rates.
Another approach is to use a two-stream encoder architecture as in~\cite{huang2015salicon}, where the image at different spatial scales is fed as input to the model, in order to extract low and high resolution information. \cite{fan2018emotional}, based on~\cite{huang2015salicon}, used a similar network adding, after feature extraction, a channel weighting subnetwork that encodes contextual information.
Differently from the above models, other works exploit adversarial training~\cite{goodfellow2014generative} for saliency prediction, such as SalGAN~\cite{pan2017salgan} and GazeGAN~\cite{8866748}.
}
Compared to saliency models for still images, saliency prediction in videos is an even more complex problem, due to the presence of the temporal dimension and to the additional computational effort it requires.
Static saliency models have been adapted to dynamic saliency by using them in frame-by-frame mode, but they are outperformed by the dynamic models that jointly process the temporal dimension. 

In recent years, a common strategy has been to extend static saliency models to the video scenario by incorporating motion features~\cite{wang2017video,shokri2020salient,sun2018sg}. For example, 
\cite{wang2017video} proposes a two-model architecture to exploit spatio-temporal features: the first module performs frame-level saliency prediction; the second module, instead, takes pairs of frames with saliency predicted by the first module, and generates a dynamic saliency map. \cite{shokri2020salient} basically employs the same architecture as~\cite{wang2017video} and self-attention, through non-local operations~\cite{wang2018non}. 
SalEMA~\cite{linardos2019simple}, instead, proposes a 2D encoder-decoder architecture with a recurrent module added to the bottleneck for integrating temporal information provided by the previous frames.
Motion cues have been also included in saliency \changeS{prediction} through either recurrent neural networks applied to spatial feature encodings or convolutional recurrent networks. OM-CNN~\cite{jiang2017predicting} is a dual-stream network that extracts spatial and temporal features using YOLO~\cite{redmon2016you} and FlowNet~\cite{dosovitskiy2015flownet}, whose respective objectness and motion features are then combined via a two-layer ConvLSTM. Similarly, ACLNet~\cite{wang2018revisiting} performs static saliency \changeS{prediction} through attention module that performs a global spatial operation on learned features. These features are then given to a ConvLSTM to model temporal information.
The recent SalSAC model~\cite{wusalsac}, leveraging the success of self-attention for saliency prediction~\cite{cornia2018predicting,wang2018revisiting}, proposes an architecture with a shuffled attention mechanism on multi-level features for better modeling of spatial saliency. Correlation features between multi-level features and shuffled attention on the same features are provided to a ConvLSTM for learning temporal cues.

With the recent availability of a large-scale saliency benchmark, i.e., DHF1K~\cite{wang2018revisiting}, 3D fully-convolutional models~\cite{bazzani2016recurrent,min2019tased}, jointly extracting spatial and temporal features, have been proposed.
RMDN~\cite{bazzani2016recurrent} processes video clips with a 3D convolutional neural network based on C3D~\cite{tran2015learning}, and then employs LSTMs to enforce temporal consistency among the segments.
TASED-Net~\cite{min2019tased} is a 3D fully-convolutional network, based on a standard encoder-decoder architecture, for video saliency detection without any additional feature processing steps.
Similarly to the above approaches, our HD$^2$S model is a 3D fully-convolutional network extending the multi-abstraction level analysis, proposed in~\cite{wang2018deep} for static saliency, to the video domain by learning spatio-temporal cues.

\changeS{Multi-level feature learning has been already applied in several application domains, most notably in object detection through the use of \emph{feature pyramid networks} (FPN)~\cite{he2020mask}. Most relevant to our approach are the works that carry out salient object detection using multi-level feature hierarchies, such as Amulet~\cite{zhang2017amulet} and DSS~\cite{hou2019deeply}. However, beside targeting static saliency prediction in images (and not in videos), those approaches apply an early-fusion mechanism of multi-level features, that are combined (through different concatenation schemas) before being further processed. Our method, instead, performs a late fusion of features: we encourage each decoding path to independently extract information from a certain abstraction layer, making sure that no inter-branch ``contamination'' may happen except at the very last layer, and thus pushing it to learn scale-specific and complementary saliency features.}
HD$^2$S also performs domain adaption to generalize across datasets without the need to be fine-tuned. Indeed, in all prediction tasks, shifts in train and test distributions may lead to a significant degradation of the model's performance. Trying to train a predictor capable of handling these shifts is commonly referred to as domain adaptation.
Among the different domain adaptation settings\footnote{An extensive review of domain adaptation approaches is out of the scope of this paper and can be found in~\cite{pan2009survey,wang2018deepda}}, we focus on \emph{unsupervised} domain adaptation, which is the task of aligning features extracted from the model across source and target domains, without any labelled samples from the latter. Several techniques have been proposed (though not for saliency prediction), such as regularizing the maximum mean discrepancy \cite{long2015learning}, minimizing correlation~\cite{sun2016deep}, or adversarial discriminator accuracy~\cite{ganin2016domain,tzeng2017adversarial}. 
An effective approach to transfer the feature distribution from source to target domains is proposed in~\cite{ganin2016domain} through the use of \emph{gradient reversal layers}, treating domain invariance as a binary classification problem. This approach addresses domain adaptation by adversarially forcing a model to solve a given task while learning features that are non-discriminative across datasets. 
In HD$^2$S we apply this strategy on multi-level features (unlike typical single-branch usage), in order to support the generalization of the saliency prediction task to datasets for which no annotations are available during training.
While unsupervised domain adaptation has been applied to image classification~\cite{ganin2016domain,tzeng2017adversarial}, face recognition~\cite{Kan_2015_ICCV}, object detection~\cite{tang2016large}, semantic segmentation~\cite{zhang2017curriculum} and video action recognition~\cite{li2018unsupervised} (among others), \changeS{our work is, to our knowledge, the first to deal with unsupervised domain adaptation on video saliency prediction. It is worthwhile to note that this is technically and fundamentally different from the form of domain adaptation proposed in UNISAL~\cite{droste2020unified}, that, instead, learns domain-specific parameters. This means that, at inference time,  UNISAL requires to know the source dataset of a given input in order to select domain-specific learned parameters. Our approach, instead, is domain-agnostic as it employs the learned parameters on any tested domain. It is also different from unsupervised salient object detection~\cite{zhang2018deep}, which, instead, attempts to predict saliency by exploiting large unlabelled or weakly-labelled samples. However, we also provide \model with domain-specific learning capabilities as in~\cite{droste2020unified}, showing how this mechanism improves performance but cannot be applied in unsupervised domain adaptation scenarios.}

\section{Method}
\subsection{Architecture overview}
\begin{figure*}[ht!]   
\centering
    \includegraphics[max width =\linewidth]{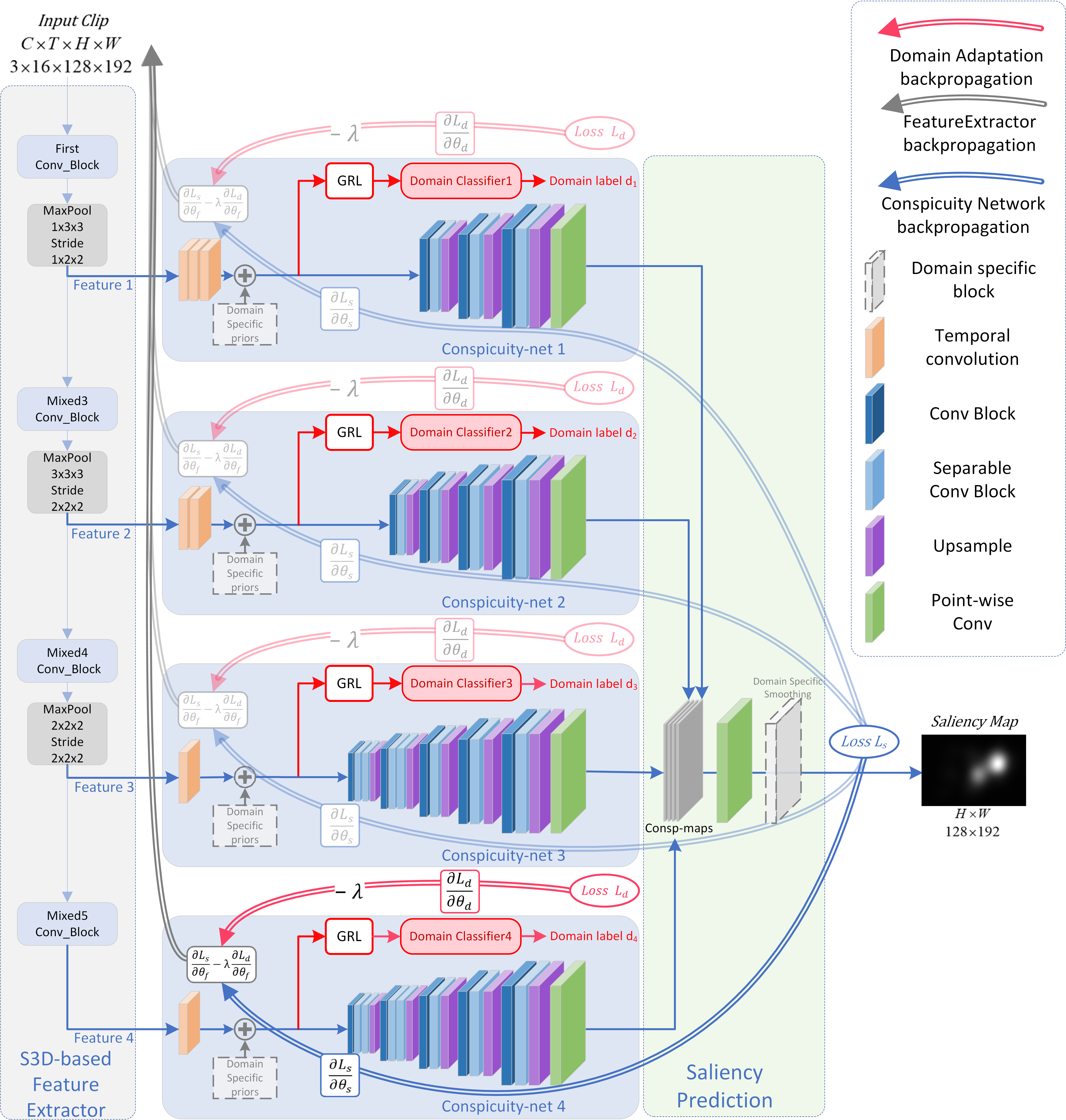}
    \caption{\changeS{\textbf{HD\textsuperscript{2}S architecture}: 
    Our multi-branch decoder predicts four \emph{conspicuity maps} at different feature abstraction levels, which are then integrated into the final saliency prediction, on which KL-divergence loss $\mathcal{L}_s$ is minimized. 
    As for \emph{unsupervised domain adaption}, each decoder branch is equipped with a gradient reversal layer (see red items) that encourages the model to learn features that generalize to a target data domain in an unsupervised way, by maximizing the classification error $\mathcal{L}_d$ on the prediction of an input sample's domain.
    Finally, \model is also provided with \emph{domain-specific} priors added to encoded features, with removed temporal dimension, and domain-specific smoothing as a last final layer.
    }}
    \label{fig:HD$^2$S}
\end{figure*}

The proposed architecture is a fully-convolutional multi-branch encoder-decoder network for saliency prediction, illustrated in Fig.~\ref{fig:HD$^2$S}. An input sequence of consecutive video frames is first processed by a \emph{feature extraction} path, which computes spatio-temporal features at different scales and abstraction levels. The extracted features serve as input to separate network branches that estimate a set of \emph{conspicuity maps} at the corresponding points in the model, while at the same time providing skip paths to ease gradient flow during training. At the output of the model, conspicuity maps are combined to predict the saliency map for the last frame in the input sequence.
Our model is trained in a supervised way on a \emph{source} dataset, for which saliency annotations are available. 

Furthermore, the base model is provided with two additional mechanisms (that can be both disabled or enabled exclusively):
\begin{itemize}
    \item \textbf{Domain adaptation} modules that aim to make the model learn, in an unsupervised way, generalizable features (see red items in Fig.~\ref{fig:HD$^2$S}).  
    In particular, each conspicuity subnetwork forks to a \emph{domain classification} path, that is trained to classify whether an input video sequence (more precisely, the corresponding features at that abstraction level) is taken from the source domain or from a \emph{target} domain, which cannot be employed for training through direct supervision since annotations are not available. 
    In order to perform this adaptation, we apply the gradient reversal technique: the feature extraction layer, shared by the conspicuity networks and the domain classifiers, is trained in an adversarial way, in order to force the model to learn features that are both discriminative and predictive --- saliency-wise --- as well as domain-invariant, in order to achieve satisfactory results even on the target domain.
    \item \textbf{Domain-specific learning} mechanism that learns specific parameters to enhance the prediction on a given dataset. 
    More specifically, we add modules (shown as light gray items in Fig.~\ref{fig:HD$^2$S}), used in a multi-source training scenario (i.e., when using in training multiple datasets at the same time), whose parameters are optimized on each individual dataset. These modules aim to modulate features shared across multiple datasets based on the test data domain and include: domain-specific priors, batch-normalization and prediction smoothing. 
\end{itemize}

At inference time, saliency maps are predicted for each frame by applying the model in a sliding window fashion, as in \cite{min2019tased}; the saliency map $\mathbf{S}_t$ at time $t$ is predicted from a sequence $\mathbf{V}_t = \left\{ \mathbf{I}_{t-T+1}, \dots, \mathbf{I}_t \right\}$, where $\mathbf{I}_t$ is the video frame at time $t$. To predict the first $T-1$ frames, we reverse the chronological order of the corresponding input clips: each $\mathbf{S}_t$ for $1 \leq t \leq T-1$ is predicted from the sequence $\mathbf{V}_t = \left\{ \mathbf{I}_{t+T-1}, \dots, I_t \right\}$. As a final post-processing step, we apply a Gaussian filter ($\sigma = 5$) for smoothing the output saliency map.

In the following, we describe each of the components of our architecture.

\subsection{Feature extractor}

The employed feature extractor performs spatio-temporal encoding of an input videoclip (16 frames of size 128$\times$192), using S3D~\cite{xie2018rethinking} as a backbone.  
It then progressively reduces the dimensions of the feature maps through 3D max pooling to 2$\times$4$\times$6 (time $\times$ height $\times$ width), while increasing the number of channels to 1024. However, in order to exploit the full potential of the learned hierarchical representations, we select feature maps at different levels of the extractor, corresponding to different abstraction details, in order to build a skip architecture able to capture multi-headed saliency responses. In our implementation, we select feature maps from the S3D backbone at the output of the second, third and fourth pooling layers and at the input of the last average pooling layer.

\subsection{Conspicuity networks}

After feature encoding, we learn several \emph{conspicuity maps} from the partial information produced at different levels of the feature extraction stack through multiple decoder networks (referred as conspicuity networks in Fig.\ref{fig:HD$^2$S}).

Each conspicuity network in the model processes one of the spatio-temporal feature blocks coming from the feature extractor and returns a single-channel saliency map, encoding the conspicuity of spatial locations at that level of abstraction. In detail, the temporal dimension of the input feature maps is gradually removed, by applying a cascade of spatially point-wise convolutions (i.e., with kernel $3\times 1\times 1$ and stride $2\times 1 \times 1$) that halve the temporal dimension at each step. The number of point-wise convolutions varies for each conspicuity network, depending on the size of the input feature maps.

After that, the (now purely spatial) set of feature maps is processed by a stack of 2D convolutional layers, interleaved with bilinear upsampling blocks, each of which doubles the spatial size of the feature maps until the original resolution of each frame is recovered.

\subsection{Saliency prediction}

\changeS{The four conspicuity maps produced by the above sub-networks are finally fused to predict saliency on the last frame of the input video. The global fusion layer consists of concatenating the four maps and performing pixel-wise 1$\times$1 convolution followed by logistic activation.}

\changeS{At training time, the whole model (feature extractor, conspicuity networks and saliency predictor) is trained supervisedly on the source dataset in order to minimize the Kullback-Leibler (KL) divergence~\cite{min2019tased,huang2015salicon}, between the predicted saliency map and conspicuity maps, and the correct target. 
More formally, given the predicted output saliency map $\mathbf{S}_t$, the four conspicuity maps $\mathbf{C}_{t,i}$ with $i=1,2,3,4$ and the ground-truth map $\mathbf{G}_t$ for a given target frame, \changeS{all normalized over pixels appropriately}, our \textit{multi-level saliency loss} $\mathcal{L}_s$ is computed as follows:
\begin{equation}
 \mathcal{L}_s\left( \mathbf{S}_t, \mathbf{C}_t, \mathbf{G}_t \right) = \sum_{j=1}^4\sum_{i} {G}_{t,i} \log \frac{{G}_{t,i}}{{C}_{t,j,i}} + \sum_{i} {G}_{t,i} \log \frac{{G}_{t,i}}{{S}_{t,i}}
\end{equation}
where index $i$ iterates over all pixels, index $j$ iterates over the four conspicuity maps, $G_{t,i}$, $S_{t,i}$ and $C_{t,i,j}$ are corresponding pixels of, respectively, the ground truth map, the output saliency map and the $j$-th conspicuity map.
}

\subsection{Domain adaptation}

In addition to training the model in a supervised way on the source domain, we also encourage the feature extractor to generalize over a target domain, without any supervision. Our unsupervised domain adaptation strategy relies on the S\emph{gradient reversal layer} (GRL) approach.

In particular, we integrate domain adaptation by inserting, in all of the conspicuity subnetworks, a branch with a gradient reversal layer and a domain classifier after the temporal-dimension removal layer (see Fig.~\ref{fig:HD$^2$S}). More formally and generally, given an input video clip $\mathbf{V}_{t}$ with associate binary domain label $d \in \left\{ 0, 1 \right\}$ (source or target, respectively), we compute a set of associated domain classification losses $\left\{ \mathcal{L}_{d,1}, \dots, \mathcal{L}_{d,4} \right\}$ from 4 domain classifiers attached to the conspicuity networks. If we indicate by $\hat{d}_i$ the probability of the input being from the target domain estimated by the $i$-th classifier, the corresponding negative log-likelihood loss is defined as:
\begin{equation}
 \mathcal{L}_{d,i}\left(d, \hat{d}_i \right) = - d \log \hat{d}_i  - (1 - d) \log \left( 1 - \hat{d}_i \right)
\end{equation}

\changeS{The overall domain classification loss is simply computed as the sum of the individual contributions, since the interaction between saliency prediction and domain adaptation is controlled by the $\lambda$ hyperparameter in the gradient reversal layers. As a result, the comprehensive loss for model training with domain adaptation is the following:
\begin{equation}
 \mathcal{L} = \mathcal{L}_s + \sum_{i=1}^4 \mathcal{L}_{d,i}
\end{equation}}

During training, we alternately pass a batch of videos from the source domain and a batch of videos from the target domain: on the former, we compute and backpropagate both the saliency prediction loss $\mathcal{L}_s$ and the domain classification loss $\mathcal{L}_d$ (with target $d = 0$); on the latter, we can only compute and backpropagate the domain classification loss $\mathcal{L}_d$ (with target $d = 1$), since no saliency annotation is available on the target domain. Minimizing the domain classification loss has the effect to train the classifiers to better discriminate between the source and the target domains, while at the same time adversarially training the feature extractor (and the initial temporal-removal layers in the conspicuity networks) to produce features that confuse the classifier, and hence that are domain-independent.

Architecturally, each domain classifier consists of a stack of 1$\times$1 spatial convolutions aimed at reducing the number of features, followed by fully-connected layers, the last of which provides binary classification prediction of the input video's domain. 

\subsection{\changeS{Domain-specific learning}}
\label{sub:domain_learning}
\changeS{In certain multi-source training scenarios (e.g., as done in~\cite{droste2020unified}), one may assume that annotations are available for all employed datasets, thus enabling supervised training on all of them. When applying our saliency prediction model to this scenario, we provide it with \emph{domain-specific operations}~\cite{droste2020unified}, which address the domain shift among different datasets. Unlike the unsupervised domain adaption setting, where we attempt to unsupervisedly learn features that generalize over multiple datasets, we here explicitly tailor learned features to the specific characteristics of each dataset.\\
In practice, we adopt a set of domain-specific techniques which have demonstrated to be effective~\cite{droste2020unified}:\\
\noindent \textbf{Domain-specific priors}. \cite{droste2020unified} thoroughly analyzed multiple video saliency benchmarks, identifying the sources of data shift among them and encoding these sources into a set of Gaussian prior maps. We employ the same strategy by initializing domain priors as in~\cite{droste2020unified}, and then letting the model learn the most suitable filters to weigh the encoded spatio-temporal features depending on the input data domain. Domain priors are used to modulate the encoded features, after removing the temporal dimension (see light gray blocks in Fig.~\ref{fig:HD$^2$S}). \\
\noindent \textbf{Domain-specific smoothing}. The optimal way in which the output map should be smoothed varies between different datasets and depends mostly on how ground truth is created. To address this issue, we learn a different Gaussian kernel (i.e, with a different value of $\sigma$) for each input data domain. Unlike~\cite{droste2020unified}, our layer is parameterized by $\sigma$ only, with convolution coefficients computed accordingly to make the filter Gaussian, while~\cite{droste2020unified} initialize domain-specific convolutional filters to be Gaussian, but they may drift to non-Gaussian as the network updates its parameters. This smoothing is applied to the global saliency map (see Fig.~\ref{fig:HD$^2$S}).\\
\noindent \textbf{Domain-specific batch normalization} aims at mitigating the impact of data distribution shift on the statistics estimated by batch normalization for inference, which may become inaccurate when computed over different benchmarks. 
Thus, we learn batch normalization statistics for each dataset independently and accordingly apply them at inference time, depending on the input domain.}

\section{Experimental Results}
\subsection{Datasets} \label{sec:datasets}

\begin{figure*}[ht!]
    \centering
    \includegraphics[width=0.32\textwidth]{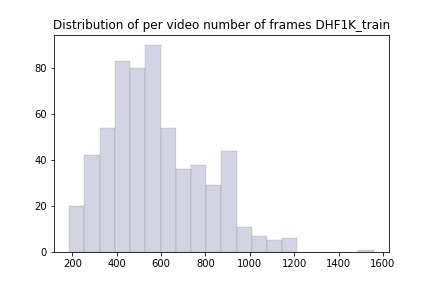}
    \includegraphics[width=0.32\textwidth]{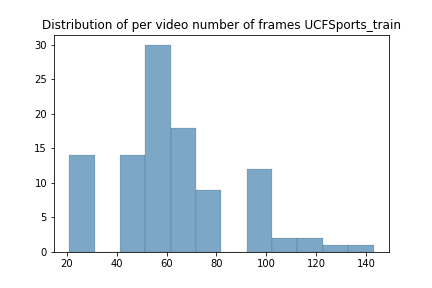}
    \includegraphics[width=0.32\textwidth]{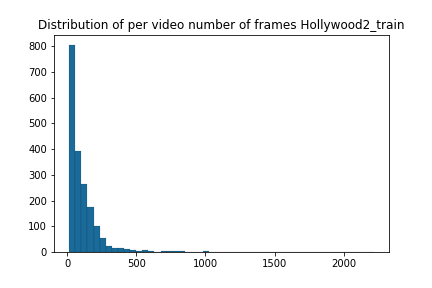}
    \caption{Statistics of the training sets of DHF1K, Hollywood and UCF Sports. 
    }
    \label{fig:dataset_overview}
\end{figure*}

\begin{itemize}
\item \textbf{DHF1K}~\cite{wang2018revisiting} consists of 1,000 high-quality videos with a large diversity of scenes, objects, types of motion, complexity of backgrounds.
\changeS{In total, it includes 582,605 frames annotated with fixation points from 17 observers during a free-viewing experiment. The dataset is split into 600/100/300 videos for training, validation and test sets, respectively.}
\changeS{The test set is not released and the results are maintained by the dataset curators\footnote{The DHF1K benchmark is available at \url{https://mmcheng.net/videosal/}}. 
}
\item \textbf{UCF Sports}~\cite{marszalek2009actions} contains 150 videos taken from the UCF Sport Action Dataset~\cite{soomro2014action}. 
Fixations are collected from 16 subjects \changeS{while attempting to identify the action that occurred in the video.}
The dataset is split into 103 videos for training, and the remaining 47 for test, for a total of around 6,500 frames for training and 3,000 frames for test. The length of the videos varies between 20 and 140 frames. 
\item \textbf{Hollywood2}~\cite{mathe2014actions} contains 6,659 video sequences and derives, like UCF Sports, from a dataset for action recognition~\cite{marszalek2009actions}. The videos are collected from 69 Hollywood movies divided into 33 training movies and 36 test movies. Similarly to UCF Sports, the annotations are collected in a task-driven way. The videos are split into 3,100 clips for training and 3,559 clips for testing. 

\item \textbf{LEDOV}~\cite{jiang2018deepvs} \changeS{includes 538 videos of daily action, sports, social activity and art performance; we employ this dataset only as a target dataset for unsupervised domain adaptation.}
\end{itemize}

Fig.~\ref{fig:dataset_overview} provides statistics on the training splits of the datasets employed for training our model: 1) UCF Sports is the smallest one in terms of available videos and average number of frames per video, thus it seems to be unsuitable for models with high capacity as they likely overfit it; 2) Hollywood2 contains the highest number of videos but the majority has a very short number of frames (see the right histogram in Fig.~\ref{fig:dataset_overview}), thus it may disadvantage methods that model temporal cues; 3) DHF1K is the most balanced in terms of videos and number of frames per videos. 

\subsection{Training procedure} \label{sec:training_procedure}
\changeS{In our experiments, we pre-train the S3D backbone on the Kinetics-400~\cite{kay2017kinetics} dataset; backbone parameters are not frozen, so they are updated during saliency prediction training.}
After empirically testing different hyperparameter configurations in order to find the best combination, the networks are trained for 2500 iterations, using Adam as optimizer~\cite{kingma2014adam} with learning rate of $10^{-3}$. \changeS{To reduce overfitting, $L_2$ regularization is applied, with a weight decay factor of $2\times 10^{-7}$}. The $\lambda$ parameter of the gradient reversal layers during training gradually varies from 0 to 1:
\begin{equation}
    \lambda = \frac{2}{ 1+e^{-10\cdot p}} -1
\end{equation}
where $p$ linearly goes from 0 to 1 according to the formula:
\begin{equation}
    p = \frac{\text{current\_iteration}}{\text{total\_iterations}}
    \label{eq:p}
\end{equation}
\changeS{Gradually increasing $\lambda$ also acts as an additional regularizer, since it prevents the model from focusing too much on the saliency prediction objective as training goes on.}
During training, sequences of $T=16$ consecutive frames are randomly sampled from the dataset's videos, and each frame is spatially resized to $128\times192$. 
We employed a batch size of 200, although for memory limitations we forward batches of 8 samples at each time, which accumulating gradients and updating the model's parameters every 25 such forward steps. When training with domain adaptation, we also forward a batch of samples from the source domain and one of samples from the target domain, and use them to update the domain classifier only.

\changeS{To evaluate performance, we use each dataset training/test split when available, with 10\% of the training data used as validation split. An exception is represented by DHF1K, since ground-truth annotations for the test set are not provided for blind assessment: in this case, when comparing to state-of-the-art methods (Tab.~\ref{tab:dhf1k_results}), \changeS{we report the test accuracy as computed by the dataset curators; while for ablation study (Tab.~\ref{tab:ablation} and~\ref{tab:ablationSingleContribution}) and domain adaptation analysis (Tab.~\ref{tab:unsupervisedDA} and~\ref{tab:supervised}), we employ the original validation set as test set.}}

\changeS{Validation results are used to perform model selection for inference on the test set.}
\changeS{When evaluating test performance in single-dataset experiments, the training, validation and test sets all come from the same domain.}\\
\changeS{In domain adaptation experiments (with labeled source and unlabeled target datasets), training and validation splits are from the source domain (whose annotations can be used at training time), while the test set is from either an unseen portion of the target domain or from a different dataset altogether.}
\\
\changeS{In multi-dataset experiments, we combine the training splits of DHF1K, UCF Sports and Hollywood2 datasets into a single training set; as validation set, we employ only DHF1K's validation split (because of its better balance compared to the other datasets, as mentioned in Sect.~\ref{sec:datasets}), while inference is carried out on each dataset's test split.
In this setting, in order to support domain-specific learning and correctly update domain-specific modules, each training mini-batch contains videos from a dataset at a time, alternating between datasets to deal with different dataset sizes}.

To compare the results obtained by the models, we use five commonly used evaluation metrics for video saliency prediction~\cite{bylinskii2018different}: Normalized Scanpath Saliency (NSS), Linear Correlation Coefficient (CC), Area under the Curve by Judd (AUC-J), Shuffled-AUC (s-AUC) and Similarity (SIM). Higher scores on each metric mean better performance.

\subsection{Video saliency prediction performance} \label{sec:saliency_performance}
\changeS{We first test the performance of our base model (without any form of adaptation) in the supervised scenario on the DHF1K test benchmark, to evaluate its capabilities in the video saliency prediction task. We then integrate domain adaptation by means of GRL layers (as shown in Fig.~\ref{fig:HD$^2$S}), using the LEDOV dataset as a target domain, due to its wider subject variability than Hollywood2 and UCF Sports. Finally, we compute the performance of HD$^2$S when using domain-specific learning, which is the form of adaptation that is most suitable with supervised learning settings and that can leverage all available annotated datasets (DHF1K, Hollywood2, UCF Sports).} \\
\changeS{Tab.~\ref{tab:dhf1k_results} shows the performance of our approach compared to the state of the art. HD$^2$S, without domain adaptation (referred to in Tab.~\ref{tab:dhf1k_results} simply as HD$^2$S), outperforms all state-of-the-art methods on three out of five metrics (NSS, AUC-J, CC) and ranks second-best on SIM and third-best on s-AUC. Note that this variant also outperforms UNISAL~\cite{droste2020unified}, which already employs domain-specific learning, on four out of five metrics. When we also enable domain-specific learning modules HD$^2$S (\emph{HD$^2$S\textsubscript{DSL}}), performance (especially NSS, CC and AUC-J) increases sensibly, and it outperforms UNISAL on all metrics, demonstrating better representational and specialization capabilities. When using HD$^2$S, with the hierarchical gradient reversal mechanism for domain adaptation(\emph{HD$^2$S\textsubscript{DA}}), performance slightly degrades as the model attempts to adapt the learned features to the target datasets (in this case, UCF-Sports, Hollywood2 and LEDOV). However, remarkably, despite this adaption mechanism, the model yields performance comparable with state-of-the-art ones.}

\begin{table}[h!]
\centering
\small\addtolength{\tabcolsep}{-2pt}
\begin{tabular}{lccccc}
\toprule
\multicolumn{6}{c}{\textbf{DHF1K}} \\
\toprule
& NSS & CC & SIM & AUC-J & s-AUC\\
\midrule
GBVS~\cite{harel2007graph} & 1.474 & 0.283 & 0.186 & 0.828 & 0.554\\
STSConvNet~\cite{bak2017spatio} &  1.632&0.325&0.197&0.834&0.581\\
Deep Net~\cite{pan2016shallow} &  1.775&0.331&0.201&0.855&0.592\\
SALICON~\cite{huang2015salicon} & 1.901&0.327&0.232&0.857&0.590\\
OM-CNN~\cite{jiang2017predicting} &  1.911&0.344&0.256&0.856&0.583\\
DVA~\cite{wang2018deep}  & 2.013&0.358&0.262&0.860&0.595\\
SalGAN~\cite{pan2017salgan} & 2.043&0.370&0.262&0.866&\textit{0.709}\\
ACLNet~\cite{wang2018revisiting} & 2.354&0.434&0.315&0.890&0.601\\
SalEMA~\cite{linardos2019simple} & 2.574	& 0.449& \textbf{0.466}& 0.890&0.667\\
STRA-Net~\cite{lai2019video} & 2.558 & 0.458& 0.355& 0.895&0.663\\
TASED-Net~\cite{min2019tased}&2.667&0.470&0.361&0.895&\textbf{0.712}\\
SalSAC~\cite{wusalsac}& 2.673&0.479 & 0.357 & 0.896&0.697\\
UNISAL~\cite{droste2020unified}& 2.776 & 0.490 & 0.390 & \textit{0.901} & 0.691\\
\midrule
\textbf{HD$^2$S} & \textit{2.781} & \textit{0.497} & \textit{0.406} & \textit{0.901} & {0.699} \\
\textbf{HD$^2$S\textsubscript{DA}} & 2.709 & 0.491 & 0.381 & 0.902 & \textit{0.709}     \\
\textbf{HD$^2$S\textsubscript{DSL}} & \textbf{2.812} & \textbf{0.503} & \textit{0.406} & \textbf{0.908} & 0.702\\
\bottomrule
\end{tabular}
\caption{Comparison of HD$^2$S, with domain adaptation (\emph{HD$^2$S\textsubscript{DA}}) and with domain-specific learning (\emph{HD$^2$S\textsubscript{DSL}}), with other state-of-the-art methods on the DHF1K test set. Our variant with domain-specific learning outperforms all state-of-the-art methods on three out of five metrics (NSS, CC, AUC-J), while ranking second-best on SIM and s-AUC.}
\label{tab:dhf1k_results}
\end{table}
\begin{table}[h!]
    \centering
    \small\addtolength{\tabcolsep}{-2pt}
    \begin{tabular}{lccccc}
        \toprule
        \multicolumn{6}{c}{\textbf{Hollywood2}} \\
        \toprule
        \textbf{Method} & \textbf{NSS} & \textbf{CC} & \textbf{SIM} & \textbf{AUC-J} & \textbf{s-AUC}\\ 
        \hline
        SALICON & 2.013 & 0.452 & 0.321 & 0.856 & 0.711\\
        DVA & 2.459 & 0.482 & 0.372 & 0.886 & 0.727\\
        ACLNet & 3.086 & 0.623 & 0.542 & 0.913 & 0.757\\
        SalEMA & 3.186 & 0.613 & 0.487 & 0.919 & 0.708\\
        STRA-Net & \textit{3.478} & 0.662 & 0.536 & 0.923 & 0.774\\
        TASED-Net & 3.302 & 0.646 & 0.507 & 0.918 & 0.768\\
        SalSAC & 3.356 & \textit{0.670} & 0.529 & 0.931 & 0.712\\
        UNISAL & \textbf{3.901} & \textbf{0.673} & 0.542 & \textit{0.934} & 0.795\\
        \midrule
        \textbf{HD$^2$S} & 3.426 & 0.668 & \textbf{0.558} & 0.927 & \textit{0.797}\\
        \textbf{HD$^2$S\textsubscript{DA}} & 3.139 & 0.653 & 0.520 & 0.927 & 0.774\\
        \textbf{HD$^2$S\textsubscript{DSL}} & 3.352 & \textit{0.670} & \textit{0.551} & \textbf{0.936} & \textbf{0.807} \\
        \toprule
        \multicolumn{6}{c}{\textbf{UCF Sports}} \\
        \toprule
        \textbf{Method} & \textbf{NSS} & \textbf{CC} & \textbf{SIM} & \textbf{AUC-J} & \textbf{s-AUC}\\
        \hline
        SALICON & 1.838 & 0.375 & 0.304 & 0.848 & 0.738\\
        DVA & 2.311 & 	0.439 & 0.339 & 0.872 & 0.725\\
        ACLNet & 2.567 & 0.510 & 0.406 & 0.897 & 0.744\\
        SalEMA & 2.638 & 0.544 & 0.431 & 0.906 & 0.740\\
        STRA-Net & 3.018 & 	0.593 & 0.479 & 0.910 & 0.751\\
        TASED-Net & 2.920 & 0.582 & 0.469 & 0.899 & 0.752\\
        SalSAC & \textbf{3.523} & \textbf{0.671} & \textbf{0.534} & \textbf{0.926} &\textbf{0.806}\\
        UNISAL & \textit{3.381} & \textit{0.644} & \textit{0.523} & \textit{0.918} & \textit{0.775}\\
        \midrule
        \textbf{HD$^2$S} & 3.001 & 0.594 & 0.493 & 0.913 & 0.773 \\
        \textbf{HD$^2$S\textsubscript{DA}} & 2.756 & 0.579 & 0.478 & 0.905 & 0.759     \\
        \textbf{HD$^2$S\textsubscript{DSL}} & 3.114 & 0.604 & 0.507 & 0.904 & 0.768\\
        \bottomrule
    \end{tabular}
    \caption{Comparison of HD\textsuperscript{2}S and its variants (with domain adaptation: HD$^2$S\textsubscript{DA};  with domain-specific learning: HD$^2$S\textsubscript{DSL}) with other state-of-the-art methods on Hollywood2 and UCF Sports datasets. In bold the best results, in italic the second-best results. }
    \label{tab:Hollywood_UCF_test}
\end{table}
\changeS{Comparing HD$^2$S with TASED-Net~\cite{min2019tased}, which also employs S3D~\cite{xie2018rethinking} as backbone, it is possible to notice that our method (with and without adaptation) significantly outperforms TASED-Net in four out of five metrics using only half of the frames employed by TASED-Net (16 versus 32). TASED-Net slightly outperforms HD$^2$S on s-AUC only, a metric 
that measures performance at the peripheral areas of the image, where a larger temporal context may allow to better capture the motion of an object. The generally better performance obtained by our method w.r.t TASED-Net demonstrates the importance of hierarchical feature learning, with equal backbone features.}
\changeS{While our model yields the highest video saliency performance on DHF1K, and performance comparable to the state of the art on Hollywood2, its performance on UCF Sports is lower than UNISAL~\cite{droste2020unified} and SalSAC~\cite{wusalsac}, as reported in Table~\ref{tab:Hollywood_UCF_test}. 
This is explained first with the smaller size of UCF Sports w.r.t. DHF1K and Hollywood2. Indeed, during training, although we use all three datasets, UCF Sports accounts to about 1\% of the total number of training video frames (DHF1K: 62\%, Hollywood2: 37\%, UCF Sports 1\%). This imbalance causes the model to overfit UCF Sports. }
\begin{figure}
    \centering
    \includegraphics[width = 0.5\textwidth]{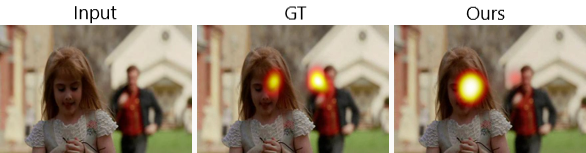}
    \caption{An example of failure, taken from Hollywood2. Despite a good prediction, \model misses to match the ground truth, as it is collected in a task-driven experiment (action recognition), thus highlighting more actions than salient objects.}
    \label{fig:hollywoodFailures}
\end{figure}

\changeS{However, the suitability of Hollywood2 and UCF Sports for saliency detection deserves a further  discussion. Indeed, both datasets' saliency annotations are collected in task-driven experiments (i.e., action recognition) and, as such, human observers tend to mainly observe specific actions rather than focusing on the salient objects themselves, which defeats the very purpose of saliency detection. An example is given in Fig.~\ref{fig:hollywoodFailures} where our model fails to match the ground truth: indeed, it focuses on the girl's face at the front (correctly, as it is the most salient area), but the ground truth mostly highlights the action of the person behind the girl. Furthermore, both datasets show a huge center bias~\cite{droste2020unified} and have a rather limited variability of spatio-temporal features, especially Hollywood2, where the majority of video clips is very short in time. Analogously, UCF Sports is significantly smaller in terms of video frames, making it hard to train 3D convolutional models (or deep learning models in general).
For all above reasons, we believe that both Hollywood2 and UCF Sports should not be used for saliency prediction.}

\section{Ablation Studies}
\changeS{To validate the importance and effectiveness of the HD\textsuperscript{2}S architectural design choices, we test some model variants (without any domain adaptation or domain-specific learning) on the validation set of the DHF1K:} 
\begin{enumerate}
    \item We first investigate the performance of our network, adding the different conspicuity nets one at a time;
    \item \changeS{We quantitatively and qualitatively evaluate the individual contribution of each conspicuity net, testing them in simple encoder-decoder architecture.}
\end{enumerate}
For the ablation study, we define as \textit{Baseline} our network in a simple encoder-decoder configuration, i.e., without the intermediate conspicuity maps and multi-level loss. More specifically, in the baseline model, the feature extractor remains unchanged, but only the deepest decoder branch (\emph{Conspicuity-net 4}) is used. 

The model variants and their performance are reported in Table \ref{tab:ablation}. The results show that: a) each conspicuity net makes its own contribution to improving the final performance; c) multi-level loss on conspicuity maps enhances saliency prediction too. Overall, these results clearly verify the effectiveness of all important design features in HD$^2$S.

\changeS{In our control experiments, we also evaluate the individual contribution of each conspicituity net by testing the performance of the model when the other decoder streams are ablated.  For example, when testing the contribution of the first conspicuity map, we use only Feature 1 (see Fig.~\ref{fig:HD$^2$S}) from the encoder stream and the related decoder stream (Conspicuity-net 1 in Fig.~\ref{fig:HD$^2$S}) and so on for the other conspicuity nets. 
Results, reported in Table~\ref{tab:ablationSingleContribution}, indicate that individually the third conspicuity net performs better than the others.}

\begin{table}[ht!] 
    \centering
    \small\addtolength{\tabcolsep}{-4pt}
        \begin{tabular}{lccccc}
            \toprule
            \textbf{} & \textbf{NSS} & \textbf{CC} & \textbf{SIM} & \textbf{AUC-J} & \textbf{s-AUC}  \\ 
            \midrule
            Consp-net4(Baseline)& 2.602 & 0.468 & 0.355 & 0.902 &  0.697\\
            \hskip 0.5em + Consp-net3 & 2.612 & 0.474 & 0.373 & 0.897 &  0.706\\
            \hskip 1em + Consp-net2 & 2.699 & 0.482 & 0.374 & 0.901  & 0.706\\
            \hskip 1.5em + Consp-net1 & 2.743 & \textbf{0.491} & 0.378 & \textbf{0.904}  &  0.704 \\
            +multi loss (\textbf{HD$^2$S}) & \textbf{2.806} & 0.489 & \textbf{0.403}  & \textbf{0.904}  & \textbf{0.705}\\
             \bottomrule
        \end{tabular}
    \caption{\textbf{Comparison of various HD$^2$S (without DA and DSL) configurations.} The \textit{Consp-net4} configuration refers to the network in a simple encoder-decoder configuration, i.e., with Conspicuity-net 4 only. The full model includes all four conspicuity networks with multi-level loss on conspicuity and saliency maps, defined in Equation~\ref{eq:p}.
    }
    \label{tab:ablation}
\end{table}
\begin{table}[ht!]
    \centering
    \small\addtolength{\tabcolsep}{-2pt}
        \begin{tabular}{lccccc}
            \toprule
            \textbf{} & \textbf{NSS} & \textbf{CC} & \textbf{SIM} & \textbf{AUC-J} & \textbf{s-AUC}  \\ 
            \midrule
            Only Consp-net1 & 2.191 & 0.392 & 0.301 & 0.871 & 0.689  \\
            Only Consp-net2 & 2.605 & 0.461 & 0.359 & 0.893 & 0.690 \\
            Only Consp-net3 & 2.663 & 0.480 & 0.359 & 0.902 & 0.697  \\
            Only Consp-net4 & 2.602 & 0.468 & 0.355 & 0.902 & 0.697  \\
            \midrule
            \textbf{Full model} & \textbf{2.806} & 0.489 & \textbf{0.403}  & \textbf{0.904}  & \textbf{0.705}\\
             \bottomrule
        \end{tabular}
    \caption{\changeS{\textbf{Individual contribution of each conspicuity net.} Each configuration refers to a simple encoder-decoder architecture with different sets of encoded features.}}
    \label{tab:ablationSingleContribution}
\end{table}

\begin{figure}[h!]
    \centering
    \includegraphics[width =0.45 \textwidth]{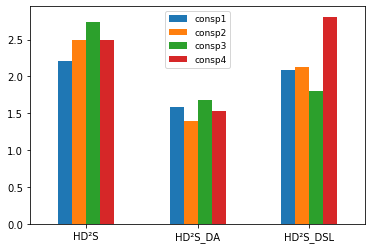}
        \caption{Weights learned by the fusion layer when integrating the four conspicuity maps on DHF1K dataset: (\textbf{left block}) Full \model model, (\textbf{middle block}) HD\textsuperscript{2}S model with domain adaptation, (\textbf{right block}) HD\textsuperscript{2}S model with domain specific learning. For the HD\textsuperscript{2}S model with domain adaptation, we use LEDOV as a target dataset.}
    \label{fig:consp_weights}
\end{figure}

\changeS{
To further elucidate this behavior, Fig.~\ref{fig:consp_weights} shows the weights learned by the fusion layer of  HD\textsuperscript{2}S model when integrating the four conspicuity maps for final prediction on the DHF1K dataset. The obtained values confirm that Conspicuity-net 3 contributes the most (see left block in Fig~\ref{fig:consp_weights}) on the prediction task for our \model model. However, it is less important when providing the model with domain-specific capabilities (which allowed it to yield the highest performance on DHF1K; see Tab.~\ref{tab:dhf1k_results}) to \model (see right blocks in Fig.~\ref{fig:consp_weights}). Furthermore, in the domain adaptation case, it can be noted how the different conspicuity maps contribute almost equally to the prediction, as a consequence of the mechanism to make the features domain-independent.}

\begin{figure*}[ht!]
    \centering
    \begin{subfigure}{0.16\textwidth}
    \includegraphics[width=\textwidth]{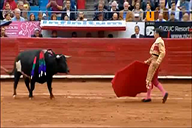}
    \label{fig:case1_input}
    \end{subfigure}
    \begin{subfigure}{0.16\linewidth}
    \includegraphics[width=\linewidth]{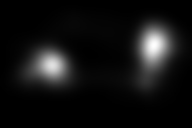}
    \label{fig:case1_map1}
    \end{subfigure}
    \begin{subfigure}{0.16\linewidth}
    \includegraphics[width=\linewidth]{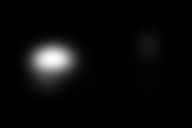}
    \label{fig:case1_map2}
    \end{subfigure}
    \begin{subfigure}{0.16\linewidth}
    \includegraphics[width=\linewidth]{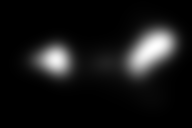}
    \label{fig:case1_map3}
    \end{subfigure}
    \begin{subfigure}{0.16\linewidth}
    \includegraphics[width=\linewidth]{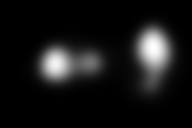}
    \label{fig:case1_map4}
    \end{subfigure}
    \begin{subfigure}{0.16\linewidth}
    \includegraphics[width=\linewidth]{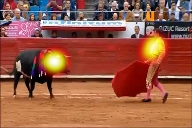}
    \label{fig:case1_overlap}
    \end{subfigure}
    \\
    \begin{subfigure}{0.16\textwidth}
    \includegraphics[width=\textwidth]{figures/607/input0795.png}
    \label{fig:case2_input}
    \end{subfigure}
    \begin{subfigure}{0.16\linewidth}
    \includegraphics[width=\linewidth]{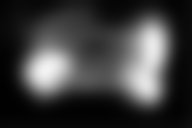}
    \label{fig:case2_map1}
    \end{subfigure}
    \begin{subfigure}{0.16\linewidth}
    \includegraphics[width=\linewidth]{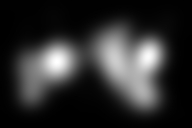}
    \label{fig:case2_map2}
    \end{subfigure}
    \begin{subfigure}{0.16\linewidth}
    \includegraphics[width=\linewidth]{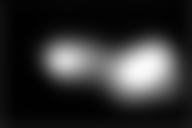}
    \label{fig:case2_map3}
    \end{subfigure}
    \begin{subfigure}{0.16\linewidth}
    \includegraphics[width=\linewidth]{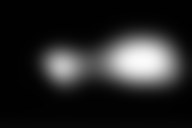}
    \label{fig:case2_map4}
    \end{subfigure}
    \begin{subfigure}{0.16\linewidth}
    \includegraphics[width=\linewidth]{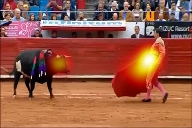}
    \label{fig:case2_overlap}
    \end{subfigure}
    \\
    \begin{subfigure}{0.16\textwidth}
    \includegraphics[width=\textwidth]{figures/607/input0795.png}
    \caption{Input frame}
    \label{fig:case3_input}
    \end{subfigure}
    \begin{subfigure}{0.16\linewidth}
    \includegraphics[width=\linewidth]{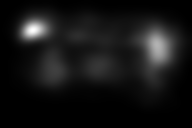}
    \caption{\texttt{Map 1}}
    \label{fig:case3_map1}
    \end{subfigure}
    \begin{subfigure}{0.16\linewidth}
    \includegraphics[width=\linewidth]{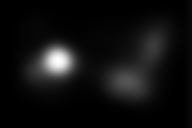}
    \caption{\texttt{Map 2}}
    \label{fig:case3_map2}
    \end{subfigure}
    \begin{subfigure}{0.16\linewidth}
    \includegraphics[width=\linewidth]{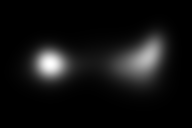}
    \caption{\texttt{Map 3}}
    \label{fig:case3_map3}
    \end{subfigure}
    \begin{subfigure}{0.16\linewidth}
    \includegraphics[width=\linewidth]{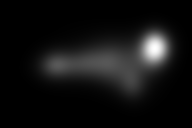}
    \caption{\texttt{Map 4}}
    \label{fig:case3_map4}
    \end{subfigure}
    \begin{subfigure}{0.16\linewidth}
    \includegraphics[width=\linewidth]{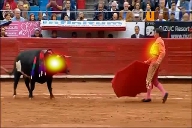}
    \caption{Final prediction}
    \label{fig:case3_overlap}
    \end{subfigure}
    \caption{Qualitative interpretation of the contribution of hierarchical decoding used under different settings. (\textbf{Top line}) HD$^2$S, (\textbf{Middle line})  HD$^2$S with domain adaptation and (\textbf{Bottom Line}) HD$^2$S with domain-specific learning. }
    \label{fig:conspicuity_examples}
\end{figure*}

\changeS{A qualitative interpretation of this behavior and on the contribution of each conspicuity map in the hierarchy is shown in Fig.~\ref{fig:conspicuity_examples}. 
When comparing the behaviour of the different decoder branches on the standard, domain adaption, and domain-specific learning regimes, the following considerations can be drawn: 1) in standard training case (top line in Fig.~\ref{fig:conspicuity_examples}), \texttt{Map 4} does not provide additional information w.r.t. \texttt{Map 3}; 2) 
in the domain adaptation scenario (middle line in Fig.~\ref{fig:conspicuity_examples}), all feature maps appear to contribute equally; 3) in the domain specific learning case (bottom line in Fig.~\ref{fig:conspicuity_examples}), \texttt{Map 4} provides additional (motion) information to \texttt{Map 3}, while on the standard learning approach the two maps encode similar information. This provides an interpretation to the parameters learned by the fusion layers, reported in Fig.~\ref{fig:consp_weights}.
Analyzing the intermediate maps in the domain specific learning (bottom line in Fig.~\ref{fig:conspicuity_examples}), we can observe that the four intermediate maps encode saliency at different levels of detail: \texttt{Map 1} extracts small background motion, \texttt{Map 2} focuses mainly on the bull, \texttt{Map 3} starts highlighting the bullfighter and, finally, \texttt{Map 4} puts more emphasis on the bullfighter. A standard encoder-decoder architecture would instead use only the last map for saliency, thus missing the bull. This highlights the usefulness of the proposed hierarchical decoding scheme.}

\section{Domain adaptation performance}
\changeS{When testing domain adaption performance, we distinguish two cases: a) the capabilities of the model to address domain shift issues, i.e., the case of reducing the shift between training and test data; and 2) the capabilities of the model to learn generalizable features that can be employed, without any additional tuning.}\\
\\
\noindent {\bf \changeS{Domain-shift.}} \changeS{To assess the performance of our hierarchical domain adaptation approach in tackling the problem of domain shift, we run a set of experiments by selecting different combinations of datasets to be employed as source domain (used in a supervised way during training) and target domain, used in an unsupervised way during training; as test set, an unseen portion from the target domain is used.
The assumption in these experiments is to perform unsupervised learning on the test domain through our hierarchical gradient reversal approach before testing on it (on the appropriate test split not used for unsupervised learning). } 

\changeS{In particular, we compare the performance of our base model in the three scenarios:} 
\begin{itemize}
    \item \changeS{\emph{Domain generalization}, i.e., the model trained supervisedly on the source domain and directly tested on the target domain, with no additional information on the test dataset used during training;
    \item \emph{Domain adaptation}, i.e., the model trained  with unsupervised adaptation on the target domain, enabled through the hierarchy of GRL layers as in our full model in Fig.~\ref{fig:HD$^2$S}};
    \item \changeS{\emph{Transfer learning}, i.e., the model (with gradient reversal disabled) trained on the source dataset and then fine-tuned (in a supervised way) on the target dataset. This scenario represents the upper bound of the evaluation and is, of course, out of the scope of pure domain adaptation, since target domain labels are available at training time.}
\end{itemize}

\changeS{Tab.~\ref{tab:unsupervisedDA} shows the results for different combinations of source and target domains. 
Two main patterns of results can be identified, depending on whether DHF1K is employed as source domain or not. In the former case \changeS{(top block of Tab.~\ref{tab:unsupervisedDA})}, it can be noticed that the employment of gradient reversal layers improves performance over all target datasets, compared to simply training on the source dataset.
When instead DHF1K is employed as target domain \changeS{(second and third blocks in Tab.~\ref{tab:unsupervisedDA})}, the use of gradient reversal layers degrades performance. This may be due to the specific characteristics of Hollywood2 and UCF Sports, which were collected in a task-driven experiments while DHF1K in a free-viewing one. Furthermore, the limited variability of spatio-temporal features from videos in Hollywood2, as shown in Fig.~\ref{fig:dataset_overview}, makes harder for the model to move clustered features and to learn more general representations. Similarly, when UCF Sports is used as source domain, the small size of the dataset makes it easier for the model to focus on the supervised saliency prediction task (on which it can easily achieve a low training loss), rather than minimizing the domain adaptation loss.
Overall, as expected, the highest performance are obtained in the transfer learning regime.}\\

\noindent \changeS{ {\bf Learning generalizable features.} We also test the capabilities of the model to learn general features by using, in the domain adaptation stream, a target dataset different from the one used for test. We specifically compute performance when training on DHF1K, adapting the learned features to LEDOV, and testing on never seen datasets (UCF and Hollywood2). Performance are reported in Table~\ref{tab:dhf1k-ledov}, which reports how the performance gain of HD$^2$S, when empowered with hierarchical gradient reversal modules, is higher than in the case of domain-shift experiments (see Table~\ref{tab:unsupervisedDA}). This demonstrates that our hierarchical domain adaptation mechanism is better at learning salience features that generalize well on multiple data domains than at addressing the domain-shift for a given dataset. }

\begin{table*}[ht!]
        \centering
        \begin{tabular}{c|lccccc}
            \toprule
            \multicolumn{7}{c}{\textbf{Source dataset: DHF1K}}\\ 
            \midrule
            \textbf{Target dataset} &\textbf{Approach} & \textbf{NSS} & \textbf{CC} & \textbf{SIM} & \textbf{AUC-J} & \textbf{s-AUC} \\
            \midrule
            \multirow{3}{*}{UCF Sports} &Domain Generalization  &  2.483 & 0.537 & 0.442 & 0.890 & 0.744 \\
            &Domain Adaptation  & 2.514 &  0.539    & 0.448 &  0.893 & 0.750 \\
            &Transfer Learning  &\textbf{ 3.001} &  \textbf{0.594}   & \textbf{0.493}    &  \textbf{0.913} & \textbf{0.773} \\
            \midrule
            \multirow{3}{*}{Hollywood2} &Domain Generalization  &  3.063 & 0.625 & 0.505 & 0.925 & 0.779 \\
            &Domain Adaptation  & 3.101 &  0.632   & 0.510 &  0.925 & 0.785 \\
            &Transfer Learning  & \textbf{3.426}    &  \textbf{0.668}    &  \textbf{0.558}    &  \textbf{0.927} & \textbf{0.797} \\
            \midrule
            \multicolumn{7}{c}{} \\
            \toprule
            \multicolumn{7}{c}{\textbf{Source dataset: UCF Sports}}\\ 
            \midrule
            \textbf{Target dataset} &\textbf{Approach} & \textbf{NSS} & \textbf{CC} & \textbf{SIM} & \textbf{AUC-J} & \textbf{s-AUC} \\
            \midrule
            \multirow{3}{*}{DHF1K} &Domain Generalization  &  2.237 & 0.405 & 0.325 & 0.880 & 0.658 \\
            &Domain Adaptation  & 2.160 &  0.392    & 0.309 &  0.877 & 0.658 \\
            &Transfer Learning  & \textbf{2.688}    &  \textbf{0.477}    & \textbf{0.374}   &  \textbf{0.896} & \textbf{0.700} \\
            \midrule
            \multirow{3}{*}{Hollywood2} &Domain Generalization  &  2.469 & 0.517 & 0.433 & 0.899 & 0.727 \\
            &Domain Adaptation  & 2.386 &  0.503   & 0.422 &  0.896 & 0.727 \\
            &Transfer Learning  & \textbf{3.298}    &  \textbf{0.657}    & \textbf{0.533}    &  \textbf{0.925} & \textbf{0.794} \\
            \midrule
            \multicolumn{7}{c}{} \\
            \toprule
            \multicolumn{7}{c}{\textbf{Source dataset: Hollywood2}}\\ 
            \midrule
            \textbf{Target dataset} &\textbf{Approach} & \textbf{NSS} & \textbf{CC} & \textbf{SIM} & \textbf{AUC-J} & \textbf{s-AUC} \\
            \midrule
            \multirow{3}{*}{DHF1K} &Domain Generalization  &  2.467 & 0.445 & 0.338 & 0.893 & 0.690 \\
            &Domain Adaptation  & 2.461 &  0.447    & 0.338 &  0.894 & 0.696 \\
            &Transfer Learning  & \textbf{2.753}    &  \textbf{0.487}    & \textbf{0.384}    & \textbf{0.898} & \textbf{0.697} \\
            \midrule
            \multirow{3}{*}{UCF Sports} &Domain Generalization  &  2.476 & 0.538 & 0.442 & 0.885 & 0.756 \\
            &Domain Adaptation  & 2.389 &  0.522   & 0.431 &  0.882 & 0.746 \\
            &Transfer Learning  & \textbf{2.780}    & \textbf{0.576}    & \textbf{0.486}    & \textbf{0.887} & \textbf{0.762} \\
            \bottomrule
        \end{tabular}
        \caption{\textbf{Analysis of domain-shift capabilities.} Performance evaluation in the \emph{domain generalization} (supervised training on source; testing on target) and \emph{domain adaptation} (supervised training on source; unsupervised training and test on target) scenarios. Upper-bound performance is measured by the \emph{transfer learning} scenario (supervised training on source and fine-tuning on target).} 
        \label{tab:unsupervisedDA}
    \end{table*}

\begin{table}[ht!]
    \centering
    \small\addtolength{\tabcolsep}{-2pt}
    \begin{tabular}{clcccc}
    \toprule
    \multicolumn{6}{c}{\textbf{source: DHF1K -  target: LEDOV }}\\
    \midrule
    Test & \textbf{Setting} & \textbf{NSS} & \textbf{CC} & \textbf{SIM} & \textbf{AUC-J} \\
    \midrule
    \multirow{2}{*}{UCF} & Generaliz.   & 2.494 & 0.536 & 0.442  & 0.889            \\
    & Adaptation   &  \textbf{2.584}  &  \textbf{0.555}  &  \textbf{0.452} & \textbf{0.900}      \\
    \midrule
    \multirow{2}{*}{Hollywood2} & Generaliz.               & 3.011            & 0.622            & 0.502 & 0.922 \\
    & Adaptation     & \textbf{3.066}   & \textbf{0.623} &  \textbf{0.505}  & \textbf{0.926}   \\
    \bottomrule
    \end{tabular}
    \caption{\textbf{Analysis of generalization capabilities.} Domain adaptation performance, respectively, with source: DHF1K, target: LEDOV; test: UCF Sports and Hollywood2. Best results in bold.}
    \label{tab:dhf1k-ledov}
\end{table}

    \begin{table*}[ht!]
        \centering
        \begin{tabular}{c|lccccc}
            \toprule
            \multicolumn{7}{c}{\textbf{Train datasets: DHF1K, UCF Sports, Hollywood2}} \\ 
            \midrule
            \textbf{Test Dataset} &\textbf{Approach} & \textbf{NSS} & \textbf{CC} & \textbf{SIM} & \textbf{AUC-J} & \textbf{s-AUC} \\
            \midrule
            \multirow{3}{*}{DHF1K} & Single-source & 2.806 & 0.489 & 0.403 & 0.904 & 0.705 \\
            & Multi-source   & 2.811 & 0.491 & 0.403 & 0.893 & \textbf{0.708} \\
            & Domain-specific  & \textbf{2.875} &  \textbf{0.500} & \textbf{0.406} &  \textbf{0.910} & 0.707 \\
            \midrule
            \multirow{3}{*}{UCF Sports} & Single-source & 2.803 & 0.589 & 0.489 & 0.879 & 0.759 \\
            & Multi-source  &  2.922 & 0.594 & 0.498 & 0.882 & 0.767 \\
            &Domain-specific  & \textbf{3.114} &  \textbf{0.604} & \textbf{0.507} &  \textbf{0.904} & \textbf{0.768} \\
            \midrule
            \multirow{3}{*}{Hollywood2} & Single-source & 3.235 & 0.660 & 0.528 & 0.919 & 0.778 \\
            &Multi-source  &  3.349 & 0.665 & 0.551 & 0.922 & 0.797 \\
            &Domain-specific  & \textbf{3.352} &  \textbf{0.670}   & \textbf{0.551} &  \textbf{0.936} & \textbf{0.807} \\
            \bottomrule
        \end{tabular}
        \caption{Performance evaluation on the multi-source and domain specific learning scenarios.}
        \label{tab:supervised}
    \end{table*}

\subsection{Multi-source training}
\changeS{A recent trend in video saliency prediction~\cite{droste2020unified} proposes \textit{multi-source training} as a means for improving performance by leveraging the larger input variability of multiple data sources. This setup also allows for the integration of \emph{domain-specific learning} capabilities, as mentioned in Sect.~\ref{sub:domain_learning}, that attempt to tune general features to specific datasets. The idea is to have a model that learns shared features across multiple datasets and then to employ domain-specific modules to adapt such features to a particular data domain.
Although these domain-specific approaches do not strictly comply with the standard unsupervised domain adaptation formulation, as they go in the exact opposite direction to learning generalizable features (since they assume that target domain labels are available at training time), it is interesting to evaluate the impact of domain-specific learning on our architecture. 
In Sect.~\ref{sec:saliency_performance} and Tab.~\ref{tab:dhf1k_results}, we already showed that the integration of domain-specific capabilities into the HD$^2$S model achieves state-of-the-art performance on DHF1K, outperforming~\cite{droste2020unified}, that introduced those techniques. Here, we complete our analysis by assessing the impact of domain-specific layers compared to multi-source domain learning. More specifically, for multi-source domain learning, we use the integration of DHF1K, Hollywood2 and UCF-Sports, as an unified dataset, for training and testing our model. As for domain specific learning, we enable the domain-specific modules (described in Sect.~\ref{sub:domain_learning}) and train their parameters using data from each individual dataset and during inference we provide, as an additional input to the model, the dataset we want to test it. We also compute performance when using single-source domain, i.e., training and test on a single dataset at a time.
The results in Tab.~\ref{tab:supervised} confirm that multi-source training by itself does not provide a much larger boost compared to single-source analysis, while domain-specific learning of dataset characteristics significantly improves performance, confirming that saliency prediction models surely benefit from embedding domain-specific layers from multiple datasets at training time.}

\subsection{Model size and runtime}
\changeS{From a computing resource perspective, Tab.~\ref{tab:proc_time} compares our model with state-of-the-art techniques, in terms of processing time and model size. Reference values for compared approaches are from~\cite{droste2020unified}.}

\begin{table}[]
    \centering
    \begin{tabular}{lcc}
    \toprule
    \textbf{Model} & \textbf{Size (MB)} & \textbf{Runtime (s)} \\
    \midrule
    Deep Net~\cite{pan2016shallow} &  103 & 0.080 \\
    SALICON~\cite{huang2015salicon} & 117 & 0.500 \\
    DVA~\cite{wang2018deep}  & 96 & 0.100 \\
    SalGAN~\cite{pan2017salgan} & 130 & 0.020 \\
    ACLNet~\cite{wang2018revisiting} & 250 & 0.020 \\
    SalEMA~\cite{linardos2019simple} & 364 & 0.010 \\
    STRA-Net~\cite{lai2019video} & 641 & 0.020 \\
    TASED-Net~\cite{min2019tased} & 82 & 0.060 \\
    UNISAL~\cite{droste2020unified} & \textbf{16} & \textbf{0.009} \\
    \textbf{HD\textsuperscript{2}S} & 116 & 0.027 \\
    \bottomrule
    \end{tabular}
    \caption{Size (in MB) and processing time (in seconds) for the proposed model and state-of-the-art approaches. Best values in bold.}
    \label{tab:proc_time}
\end{table}

\changeS{UNISAL is the most resource-efficient approach, in both time and space (thanks to its MobileNetV2~\cite{sandler2018mobilenetv2} backbone); our approach achieves average values on those metrics, while performing better than most in terms of prediction accuracy, as shown in the previous sections.}

\section{Qualitative analysis}
\begin{figure*}[h!]
    \centering
    \begin{subfigure}{0.49\textwidth}
        \begin{subfigure}{\textwidth}
        \rotatebox[origin=c]{90}{\changeF{$^2$}Input\changeF{$^2$}}
            \begin{subfigure}{0.3\textwidth}
                \includegraphics[width = \textwidth]{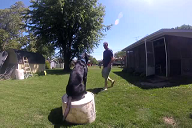}
            \end{subfigure}
            \begin{subfigure}{0.3\textwidth}
                \includegraphics[width = \textwidth]{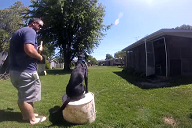}
            \end{subfigure}
            \begin{subfigure}{0.3\textwidth}
                \includegraphics[width = \textwidth]{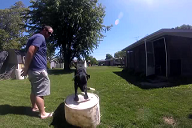}
            \end{subfigure}
        \end{subfigure}
        \begin{subfigure}{\textwidth}
        \rotatebox[origin=c]{90}{\changeF{p$^2$}GT\changeF{p$^2$}}
            \begin{subfigure}{0.3\textwidth}
                \includegraphics[width = \textwidth]{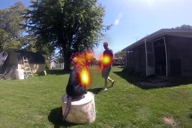}
            \end{subfigure}
            \begin{subfigure}{0.3\textwidth}
                \includegraphics[width = \textwidth]{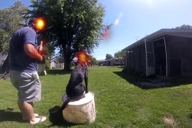}
            \end{subfigure}
            \begin{subfigure}{0.3\textwidth}
                \includegraphics[width = \textwidth]{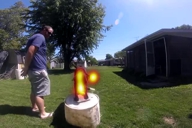}
            \end{subfigure}
        \end{subfigure}
        \begin{subfigure}{\textwidth}
        \rotatebox[origin=c]{90}{\changeF{p$^2$}\model\changeF{p$^2$}}
            \begin{subfigure}{0.3\textwidth}
                \includegraphics[width = \textwidth]{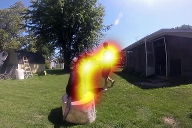}
            \end{subfigure}
            \begin{subfigure}{0.3\textwidth}
                \includegraphics[width = \textwidth]{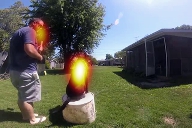}
            \end{subfigure}
            \begin{subfigure}{0.3\textwidth}
                \includegraphics[width = \textwidth]{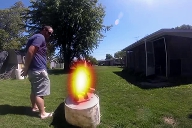}
            \end{subfigure}
        \end{subfigure}
    \end{subfigure}
    \begin{subfigure}{0.49\textwidth}
        \begin{subfigure}{\textwidth}
            \begin{subfigure}{0.3\textwidth}
                \includegraphics[width = \textwidth]{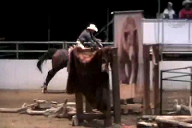}
            \end{subfigure}
            \begin{subfigure}{0.3\textwidth}
                \includegraphics[width = \textwidth]{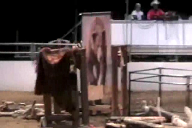}
            \end{subfigure}
            \begin{subfigure}{0.3\textwidth}
                \includegraphics[width = \textwidth]{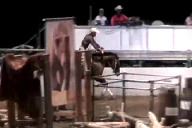}
            \end{subfigure}
        \end{subfigure}
        \begin{subfigure}{\textwidth}
            \begin{subfigure}{0.3\textwidth}
                \includegraphics[width = \textwidth]{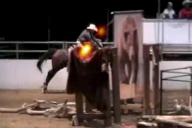}
            \end{subfigure}
            \begin{subfigure}{0.3\textwidth}
                \includegraphics[width = \textwidth]{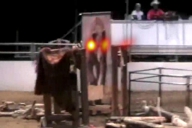}
            \end{subfigure}
            \begin{subfigure}{0.3\textwidth}
                \includegraphics[width = \textwidth]{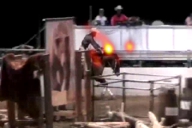}
            \end{subfigure}
        \end{subfigure}
        \begin{subfigure}{\textwidth}
            \begin{subfigure}{0.3\textwidth}
                \includegraphics[width = \textwidth]{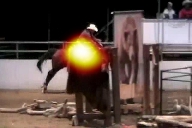}
            \end{subfigure}
            \begin{subfigure}{0.3\textwidth}
                \includegraphics[width = \textwidth]{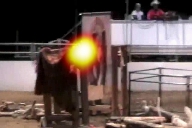}
            \end{subfigure}
            \begin{subfigure}{0.3\textwidth}
                \includegraphics[width = \textwidth]{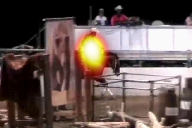}
            \end{subfigure}
        \end{subfigure}
    \end{subfigure}
    \\ \vspace{0.5em} 
    \begin{subfigure}{0.49\textwidth}
        \begin{subfigure}{\textwidth}
        \rotatebox[origin=c]{90}{\changeF{$^2$}Input\changeF{$^2$}}
            \begin{subfigure}{0.3\textwidth}
                \includegraphics[width = \textwidth]{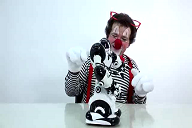}
            \end{subfigure}
            \begin{subfigure}{0.3\textwidth}
                \includegraphics[width = \textwidth]{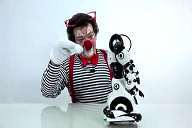}
            \end{subfigure}
            \begin{subfigure}{0.3\textwidth}
                \includegraphics[width = \textwidth]{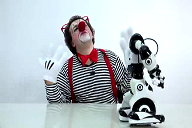}
            \end{subfigure}
        \end{subfigure}
        \begin{subfigure}{\textwidth}
        \rotatebox[origin=c]{90}{\changeF{p$^2$}GT\changeF{p$^2$}}
            \begin{subfigure}{0.3\textwidth}
                \includegraphics[width = \textwidth]{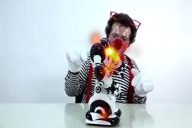}
            \end{subfigure}
            \begin{subfigure}{0.3\textwidth}
                \includegraphics[width = \textwidth]{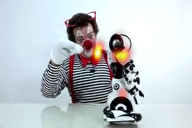}
            \end{subfigure}
            \begin{subfigure}{0.3\textwidth}
                \includegraphics[width = \textwidth]{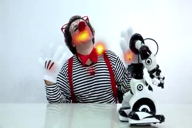}
            \end{subfigure}
        \end{subfigure}
        \begin{subfigure}{\textwidth}
        \rotatebox[origin=c]{90}{\changeF{p$^2$}\model\changeF{p$^2$}}
            \begin{subfigure}{0.3\textwidth}
                \includegraphics[width = \textwidth]{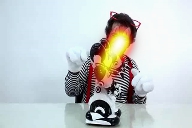}
            \end{subfigure}
            \begin{subfigure}{0.3\textwidth}
                \includegraphics[width = \textwidth]{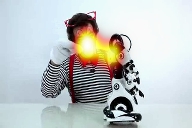}
            \end{subfigure}
            \begin{subfigure}{0.3\textwidth}
                \includegraphics[width = \textwidth]{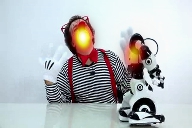}
            \end{subfigure}
        \end{subfigure}
    \end{subfigure}
    \begin{subfigure}{0.49\textwidth}
        \begin{subfigure}{\textwidth}
            \begin{subfigure}{0.3\textwidth}
                \includegraphics[width = \textwidth]{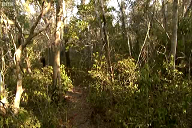}
            \end{subfigure}
            \begin{subfigure}{0.3\textwidth}
                \includegraphics[width = \textwidth]{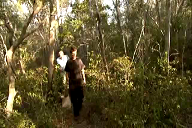}
            \end{subfigure}
            \begin{subfigure}{0.3\textwidth}
                \includegraphics[width = \textwidth]{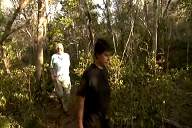}
            \end{subfigure}
        \end{subfigure}
        \begin{subfigure}{\textwidth}
            \begin{subfigure}{0.3\textwidth}
                \includegraphics[width = \textwidth]{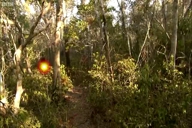}
            \end{subfigure}
            \begin{subfigure}{0.3\textwidth}
                \includegraphics[width = \textwidth]{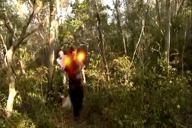}
            \end{subfigure}
            \begin{subfigure}{0.3\textwidth}
                \includegraphics[width = \textwidth]{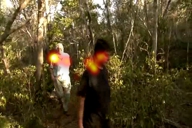}
            \end{subfigure}
        \end{subfigure}
        \begin{subfigure}{\textwidth}
            \begin{subfigure}{0.3\textwidth}
                \includegraphics[width = \textwidth]{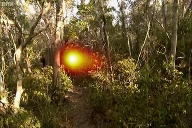}
            \end{subfigure}
            \begin{subfigure}{0.3\textwidth}
                \includegraphics[width = \textwidth]{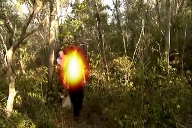}
            \end{subfigure}
            \begin{subfigure}{0.3\textwidth}
                \includegraphics[width = \textwidth]{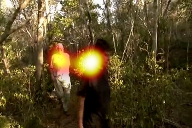}
            \end{subfigure}
        \end{subfigure}
    \end{subfigure}
    \caption{Qualitative evaluation of the proposed model on the DHF1K validation set. Comparison of the saliency predicted by our model with the ground truth on some frames: (left block) saliency with multiple objects, (upper-right block) saliency on an occluded object, (lower-right block) saliency on moving objects whose appearance changes rapidly among consecutive frames.}
    \label{fig:good_examples}
\end{figure*}
\begin{figure}[ht!]
    \centering
    \begin{subfigure}{0.5\textwidth}
        \begin{subfigure}{0.3\textwidth}
            \captionsetup{labelformat=empty}
            \caption{INPUT}
            \includegraphics[width = \textwidth]{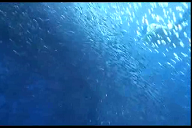}
        \end{subfigure}
        \begin{subfigure}{0.3\textwidth}
            \captionsetup{labelformat=empty}
            \caption{GT}
            \includegraphics[width = \textwidth]{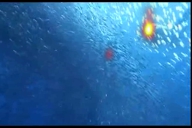}
        \end{subfigure}
        \begin{subfigure}{0.3\textwidth}
            \captionsetup{labelformat=empty}
            \caption{OURS}
            \includegraphics[width = \textwidth]{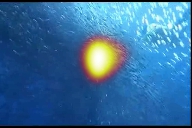}
        \end{subfigure}
    \end{subfigure}
    \begin{subfigure}{0.5\textwidth}
        \begin{subfigure}{0.3\textwidth}
            \includegraphics[width = \textwidth]{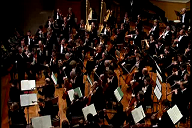}
        \end{subfigure}
        \begin{subfigure}{0.3\textwidth}
            \includegraphics[width = \textwidth]{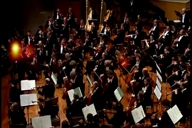}
        \end{subfigure}
        \begin{subfigure}{0.3\textwidth}
            \includegraphics[width = \textwidth]{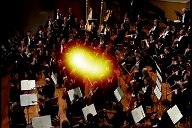}
        \end{subfigure}
    \end{subfigure}
    \begin{subfigure}{0.5\textwidth}
        \begin{subfigure}{0.3\textwidth}
            \includegraphics[width = \textwidth]{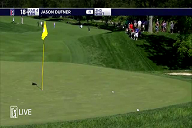}
        \end{subfigure}
        \begin{subfigure}{0.3\textwidth}
            \includegraphics[width = \textwidth]{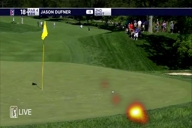}
        \end{subfigure}
        \begin{subfigure}{0.3\textwidth}
            \includegraphics[width = \textwidth]{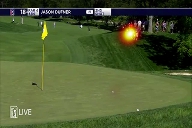}
        \end{subfigure}
    \end{subfigure}
    \caption{\changeS{Examples of failures. The model struggles with small objects and small motion: in the first two cases, the model missed the salient region and highlights a generic prior; in the third example, the model does not manage to identify the golf ball, focusing instead on a man in a red shirt, standing out from the surroundings.}}
    \label{fig:failures}
\end{figure}

\changeS{We here report quantitative analysis of the results obtained by our model. Fig.~\ref{fig:good_examples} shows examples of saliency predictions made by our HD\textsuperscript{2}S model with domain-specific learning on the DHF1K benchmark. 
The model is able to effectively face object occlusion, multiple objects, fast motion, strong camera motion, stationary objects, saliency shift, camera focus change, low-light condition. Sample videos of how our model works are also given in the GitHub page of the project.}
\changeS{Fig.~\ref{fig:failures}, instead, shows example of failures that typically happen in case of small global motion or small objects. These failures can be caused by the spatial resolution at which input images are scaled before being processed by the model (128$\times$192). Indeed, in the first two cases of Fig.~\ref{fig:failures} the models is unable to identify the correct salient region (located in a lateral region of the scene), and instead predicts a generic prior-driven center region. In the last case, the model fails to detect the movement of a golf ball towards the hole (a slow movement of a small object), and erroneously predicts as salient the upper-right region of the scene, where a man with a red shirt significantly stands out from the surroundings.}

\section{Conclusion}

\changeS{In this work, we propose HD\textsuperscript{2}S, a new fully-convolutional network for video saliency prediction. The key architectural elements of our proposed approach include a multi-branch decoder which acts at different feature abstraction layers to independently estimate \emph{conspicuity maps}, which are then combined into the final prediction, and an unsupervised domain adaptation mechanism that enables our model to learn features that, at the same time, allow it to reach state-of-the-art performance on supervised saliency prediction, while generalizing to domains for which no annotations are provided at training time. Additionally, when employing domain-specific learning techniques, as introduced in~\cite{droste2020unified}, our model's performance on the supervised saliency prediction task further improves.}

\changeS{Comparing our approach with state-of-the-art models, we find that our late-fusion mechanism of multi-level saliency features provides a significant boost to performance: our ablation studies show that the gradual integration of multiple abstraction levels positively affects prediction accuracy.  This is also confirmed by analyzing the learned weights. Interestingly, the impact of each conspicuity map (and, therefore, of each abstraction level of learned features) seems to vary depending on the employed domain adaptation mechanism: high-level features become predominant when domain-specific learning is applied (possibly due to the larger data distribution variability introduced by multi-source training, which causes shallower features to generalize less), while all conspicuity maps become similarly important when unsupervised domain adaptation is applied, which can be explained through the action of gradient reversal layers, which actively encourage features to become domain-independent and thus to be equally effective at multiple scales.}

\changeS{While the model performs well in several complex cases --- e.g., in presence of multiple objects, occlusions and appareance changes --- there are certain conditions in which we find room for improvement. Most of them involve the presence of small objects and small motion, where the model fails to correctly locate areas of interest, and the prediction is dominated by the prior.}

\changeS{These situations could benefit from working at higher resolution, given sufficient computing resources, or in a patch-based fashion, to the detriment of inference times. However, major failures seem to be related to the specific characteristics of datasets: Hollywood2 and UCF Sports, for instance, are annotated with task-driven gaze fixations, rather then free-view the scene. This, of course, negatively affects methods that instead attempt to predict bottom-up saliency. Improved dataset availability curation for video saliency prediction may be an enabling factor for the advancement in the field.}



\clearpage
\onecolumn

\begin{thebibliography}{10}

\bibitem{badrinarayanan2017segnet}
Vijay Badrinarayanan, Alex Kendall, and Roberto Cipolla.
\newblock Segnet: A deep convolutional encoder-decoder architecture for image
  segmentation.
\newblock {\em IEEE transactions on pattern analysis and machine intelligence},
  39(12):2481--2495, 2017.

\bibitem{bak2017spatio}
Cagdas Bak, Aysun Kocak, Erkut Erdem, and Aykut Erdem.
\newblock Spatio-temporal saliency networks for dynamic saliency prediction.
\newblock {\em IEEE Transactions on Multimedia}, 20(7):1688--1698, 2017.

\bibitem{bazzani2016recurrent}
Loris Bazzani, Hugo Larochelle, and Lorenzo Torresani.
\newblock Recurrent mixture density network for spatiotemporal visual
  attention.
\newblock {\em arXiv preprint arXiv:1603.08199}, 2016.

\bibitem{borji2015cat2000}
Ali Borji and Laurent Itti.
\newblock Cat2000: A large scale fixation dataset for boosting saliency
  research.
\newblock {\em arXiv preprint arXiv:1505.03581}, 2015.

\bibitem{bylinskii2018different}
Zoya Bylinskii, Tilke Judd, Aude Oliva, Antonio Torralba, and Fr{\'e}do Durand.
\newblock What do different evaluation metrics tell us about saliency models?
\newblock {\em IEEE transactions on pattern analysis and machine intelligence},
  41(3):740--757, 2018.

\bibitem{8866748}
Z.~{Che}, A.~{Borji}, G.~{Zhai}, X.~{Min}, G.~{Guo}, and P.~{Le Callet}.
\newblock How is gaze influenced by image transformations? dataset and model.
\newblock {\em IEEE Transactions on Image Processing}, 29:2287--2300, 2020.

\bibitem{chen2018saliency}
Yangyu Chen, Weigang Zhang, Shuhui Wang, Liang Li, and Qingming Huang.
\newblock Saliency-based spatiotemporal attention for video captioning.
\newblock In {\em 2018 IEEE Fourth International Conference on Multimedia Big
  Data (BigMM)}, pages 1--8. IEEE, 2018.

\bibitem{8400593}
M.~{Cornia}, L.~{Baraldi}, G.~{Serra}, and R.~{Cucchiara}.
\newblock Predicting human eye fixations via an lstm-based saliency attentive
  model.
\newblock {\em IEEE Transactions on Image Processing}, 27(10):5142--5154, 2018.

\bibitem{cornia2016deep}
Marcella Cornia, Lorenzo Baraldi, Giuseppe Serra, and Rita Cucchiara.
\newblock A deep multi-level network for saliency prediction.
\newblock In {\em 2016 23rd International Conference on Pattern Recognition
  (ICPR)}, pages 3488--3493. IEEE, 2016.

\bibitem{cornia2018predicting}
Marcella Cornia, Lorenzo Baraldi, Giuseppe Serra, and Rita Cucchiara.
\newblock Predicting human eye fixations via an lstm-based saliency attentive
  model.
\newblock {\em IEEE Transactions on Image Processing}, 27(10):5142--5154, 2018.

\bibitem{dosovitskiy2015flownet}
Alexey Dosovitskiy, Philipp Fischer, Eddy Ilg, Philip Hausser, Caner Hazirbas,
  Vladimir Golkov, Patrick Van Der~Smagt, Daniel Cremers, and Thomas Brox.
\newblock Flownet: Learning optical flow with convolutional networks.
\newblock In {\em ICCV}, pages 2758--2766, 2015.

\bibitem{droste2020unified}
Richard Droste, Jianbo Jiao, and J~Alison Noble.
\newblock Unified image and video saliency modeling.
\newblock In {\em European Conference on Computer Vision}, pages 419--435.
  Springer, 2020.

\bibitem{fan2018emotional}
Shaojing Fan, Zhiqi Shen, Ming Jiang, Bryan~L Koenig, Juan Xu, Mohan~S
  Kankanhalli, and Qi~Zhao.
\newblock Emotional attention: A study of image sentiment and visual attention.
\newblock In {\em Proceedings of the IEEE Conference on computer vision and
  pattern recognition}, pages 7521--7531, 2018.

\bibitem{ganin2016domain}
Yaroslav Ganin, Evgeniya Ustinova, Hana Ajakan, Pascal Germain, Hugo
  Larochelle, Fran{\c{c}}ois Laviolette, Mario Marchand, and Victor Lempitsky.
\newblock Domain-adversarial training of neural networks.
\newblock {\em The Journal of Machine Learning Research}, 17(1):2096--2030,
  2016.

\bibitem{Girshick_2015_ICCV}
Ross Girshick.
\newblock Fast r-cnn.
\newblock In {\em Proceedings of the IEEE International Conference on Computer
  Vision (ICCV)}, December 2015.

\bibitem{goodfellow2014generative}
Ian~J Goodfellow, Jean Pouget-Abadie, Mehdi Mirza, Bing Xu, David Warde-Farley,
  Sherjil Ozair, Aaron Courville, and Yoshua Bengio.
\newblock Generative adversarial networks.
\newblock {\em arXiv preprint arXiv:1406.2661}, 2014.

\bibitem{guraya2010predictive}
Fahad Fazal~Elahi Guraya, Faouzi~Alaya Cheikh, Alain Tremeau, Yubing Tong, and
  Hubert Konik.
\newblock Predictive saliency maps for surveillance videos.
\newblock In {\em 2010 Ninth International Symposium on Distributed Computing
  and Applications to Business, Engineering and Science}, pages 508--513. IEEE,
  2010.

\bibitem{harel2007graph}
Jonathan Harel, Christof Koch, and Pietro Perona.
\newblock Graph-based visual saliency.
\newblock In {\em NIPS}, pages 545--552, 2007.

\bibitem{he2020mask}
Kaiming He, Georgia Gkioxari, Piotr Doll{\'{a}}r, and Ross~B. Girshick.
\newblock Mask {R-CNN}.
\newblock {\em {IEEE} Trans. Pattern Anal. Mach. Intell.}, 42(2):386--397,
  2020.

\bibitem{hou2019deeply}
Qibin Hou, Ming{-}Ming Cheng, Xiaowei Hu, Ali Borji, Zhuowen Tu, and Philip
  H.~S. Torr.
\newblock Deeply supervised salient object detection with short connections.
\newblock {\em {IEEE} Trans. Pattern Anal. Mach. Intell.}, 41(4):815--828,
  2019.

\bibitem{huang2015salicon}
Xun Huang, Chengyao Shen, Xavier Boix, and Qi~Zhao.
\newblock Salicon: Reducing the semantic gap in saliency prediction by adapting
  deep neural networks.
\newblock In {\em ICCV}, pages 262--270, 2015.

\bibitem{itti1998model}
Laurent Itti, Christof Koch, and Ernst Niebur.
\newblock A model of saliency-based visual attention for rapid scene analysis.
\newblock {\em IEEE Transactions on pattern analysis and machine intelligence},
  20(11):1254--1259, 1998.

\bibitem{jia2020eml}
Sen Jia and Neil~DB Bruce.
\newblock Eml-net: An expandable multi-layer network for saliency prediction.
\newblock {\em Image and Vision Computing}, 95:103887, 2020.

\bibitem{jiang2018deepvs}
Lai Jiang, Mai Xu, Tie Liu, Minglang Qiao, and Zulin Wang.
\newblock Deepvs: A deep learning based video saliency prediction approach.
\newblock In {\em ECCV}, pages 602--617, 2018.

\bibitem{jiang2017predicting}
Lai Jiang, Mai Xu, and Zulin Wang.
\newblock Predicting video saliency with object-to-motion cnn and two-layer
  convolutional lstm.
\newblock {\em arXiv preprint arXiv:1709.06316}, 2017.

\bibitem{jiang2015salicon}
Ming Jiang, Shengsheng Huang, Juanyong Duan, and Qi~Zhao.
\newblock Salicon: Saliency in context.
\newblock In {\em Proceedings of the IEEE conference on computer vision and
  pattern recognition}, pages 1072--1080, 2015.

\bibitem{judd2012benchmark}
Tilke Judd, Fr{\'e}do Durand, and Antonio Torralba.
\newblock A benchmark of computational models of saliency to predict human
  fixations.
\newblock 2012.

\bibitem{Kan_2015_ICCV}
Meina Kan, Shiguang Shan, and Xilin Chen.
\newblock Bi-shifting auto-encoder for unsupervised domain adaptation.
\newblock In {\em ICCV}, 2015.

\bibitem{kay2017kinetics}
Will Kay, Joao Carreira, Karen Simonyan, Brian Zhang, Chloe Hillier, Sudheendra
  Vijayanarasimhan, Fabio Viola, Tim Green, Trevor Back, Paul Natsev, et~al.
\newblock The kinetics human action video dataset.
\newblock {\em arXiv preprint arXiv:1705.06950}, 2017.

\bibitem{kingma2014adam}
Diederik~P Kingma and Jimmy Ba.
\newblock Adam: A method for stochastic optimization.
\newblock {\em arXiv preprint arXiv:1412.6980}, 2014.

\bibitem{kroner2020contextual}
Alexander Kroner, Mario Senden, Kurt Driessens, and Rainer Goebel.
\newblock Contextual encoder--decoder network for visual saliency prediction.
\newblock {\em Neural Networks}, 129:261--270, 2020.

\bibitem{Kummerer_2017_ICCV}
Matthias Kummerer, Thomas S.~A. Wallis, Leon~A. Gatys, and Matthias Bethge.
\newblock Understanding low- and high-level contributions to fixation
  prediction.
\newblock In {\em Proceedings of the IEEE International Conference on Computer
  Vision (ICCV)}, Oct 2017.

\bibitem{lai2019video}
Qiuxia Lai, Wenguan Wang, Hanqiu Sun, and Jianbing Shen.
\newblock Video saliency prediction using spatiotemporal residual attentive
  networks.
\newblock {\em IEEE Transactions on Image Processing}, 29:1113--1126, 2019.

\bibitem{li2018unsupervised}
Junnan Li, Yongkang Wong, Qi~Zhao, and Mohan~S Kankanhalli.
\newblock Unsupervised learning of view-invariant action representations.
\newblock In {\em NIPS}, pages 1254--1264, 2018.

\bibitem{li2007fast}
Shan Li and MC~Lee.
\newblock Fast visual tracking using motion saliency in video.
\newblock In {\em ICASSP}, volume~1, pages I--1073. IEEE, 2007.

\bibitem{lim2014crowd}
Mei~Kuan Lim, Ven~Jyn Kok, Chen~Change Loy, and Chee~Seng Chan.
\newblock Crowd saliency detection via global similarity structure.
\newblock In {\em 2014 22nd International Conference on Pattern Recognition},
  pages 3957--3962. IEEE, 2014.

\bibitem{linardos2019simple}
Panagiotis Linardos, Eva Mohedano, Juan~Jose Nieto, Noel~E O'Connor, Xavier
  Giro-i Nieto, and Kevin McGuinness.
\newblock Simple vs complex temporal recurrences for video saliency prediction.
\newblock {\em arXiv preprint arXiv:1907.01869}, 2019.

\bibitem{liu2010learning}
Tie Liu, Zejian Yuan, Jian Sun, Jingdong Wang, Nanning Zheng, Xiaoou Tang, and
  Heung-Yeung Shum.
\newblock Learning to detect a salient object.
\newblock {\em IEEE Transactions on Pattern analysis and machine intelligence},
  33(2):353--367, 2010.

\bibitem{long2015learning}
Mingsheng Long, Yue Cao, Jianmin Wang, and Michael Jordan.
\newblock Learning transferable features with deep adaptation networks.
\newblock In {\em International conference on machine learning}, pages 97--105.
  PMLR, 2015.

\bibitem{lu2017crowd}
Li~Lu, Jia He, Zhijie Xu, Yuanping Xu, Chaolong Zhang, Jing Wang, and Jianhua
  Adu.
\newblock Crowd behavior understanding through siof feature analysis.
\newblock In {\em 2017 23rd International Conference on Automation and
  Computing (ICAC)}, pages 1--6. IEEE, 2017.

\bibitem{marszalek2009actions}
Marcin Marszalek, Ivan Laptev, and Cordelia Schmid.
\newblock Actions in context.
\newblock In {\em CVPR}, pages 2929--2936. IEEE, 2009.

\bibitem{6942210}
S.~{Mathe} and C.~{Sminchisescu}.
\newblock Actions in the eye: Dynamic gaze datasets and learnt saliency models
  for visual recognition.
\newblock {\em IEEE Transactions on Pattern Analysis and Machine Intelligence},
  37(7):1408--1424, 2015.

\bibitem{mathe2014actions}
Stefan Mathe and Cristian Sminchisescu.
\newblock Actions in the eye: Dynamic gaze datasets and learnt saliency models
  for visual recognition.
\newblock {\em IEEE transactions on pattern analysis and machine intelligence},
  37(7):1408--1424, 2014.

\bibitem{min2019tased}
Kyle Min and Jason~J Corso.
\newblock Tased-net: Temporally-aggregating spatial encoder-decoder network for
  video saliency detection.
\newblock In {\em ICCV}, pages 2394--2403, 2019.

\bibitem{nguyen2013static}
Tam~V Nguyen, Mengdi Xu, Guangyu Gao, Mohan Kankanhalli, Qi~Tian, and Shuicheng
  Yan.
\newblock Static saliency vs. dynamic saliency: a comparative study.
\newblock In {\em ACM MM}, pages 987--996, 2013.

\bibitem{noh2015learning}
Hyeonwoo Noh, Seunghoon Hong, and Bohyung Han.
\newblock Learning deconvolution network for semantic segmentation.
\newblock In {\em ICCV}, pages 1520--1528, 2015.

\bibitem{pan2017salgan}
Junting Pan, Cristian~Canton Ferrer, Kevin McGuinness, Noel~E O'Connor, Jordi
  Torres, Elisa Sayrol, and Xavier Giro-i Nieto.
\newblock Salgan: Visual saliency prediction with generative adversarial
  networks.
\newblock {\em arXiv preprint arXiv:1701.01081}, 2017.

\bibitem{pan2016shallow}
Junting Pan, Elisa Sayrol, Xavier Giro-i Nieto, Kevin McGuinness, and Noel~E
  O'Connor.
\newblock Shallow and deep convolutional networks for saliency prediction.
\newblock In {\em CVPR}, pages 598--606, 2016.

\bibitem{pan2009survey}
Sinno~Jialin Pan and Qiang Yang.
\newblock A survey on transfer learning.
\newblock {\em IEEE Transactions on knowledge and data engineering},
  22(10):1345--1359, 2009.

\bibitem{redmon2016you}
Joseph Redmon, Santosh Divvala, Ross Girshick, and Ali Farhadi.
\newblock You only look once: Unified, real-time object detection.
\newblock In {\em CVPR}, pages 779--788, 2016.

\bibitem{ronneberger2015u}
Olaf Ronneberger, Philipp Fischer, and Thomas Brox.
\newblock U-net: Convolutional networks for biomedical image segmentation.
\newblock In {\em MICCAI}, pages 234--241. Springer, 2015.

\bibitem{sandler2018mobilenetv2}
Mark Sandler, Andrew Howard, Menglong Zhu, Andrey Zhmoginov, and Liang-Chieh
  Chen.
\newblock Mobilenetv2: Inverted residuals and linear bottlenecks.
\newblock In {\em CVPR}, pages 4510--4520, 2018.

\bibitem{shao2005tracking}
Jie Shao, Shaohua~Kevin Zhou, and Rama Chellappa.
\newblock Tracking algorithm using background- foreground motion models and
  multiple cues [surveillance video applications].
\newblock In {\em ICASSP}, volume~2, pages ii--233. IEEE, 2005.

\bibitem{shokri2020salient}
Mohammad Shokri, Ahad Harati, and Kimya Taba.
\newblock Salient object detection in video using deep non-local neural
  networks.
\newblock {\em Journal of Visual Communication and Image Representation},
  68:102769, 2020.

\bibitem{soomro2014action}
Khurram Soomro and Amir~R Zamir.
\newblock Action recognition in realistic sports videos.
\newblock In {\em Computer vision in sports}, pages 181--208. Springer, 2014.

\bibitem{sun2016deep}
Baochen Sun and Kate Saenko.
\newblock Deep coral: Correlation alignment for deep domain adaptation.
\newblock In {\em ECCV}, pages 443--450. Springer, 2016.

\bibitem{sun2018sg}
Meijun Sun, Ziqi Zhou, Qinghua Hu, Zheng Wang, and Jianmin Jiang.
\newblock Sg-fcn: A motion and memory-based deep learning model for video
  saliency detection.
\newblock {\em IEEE transactions on cybernetics}, 49(8):2900--2911, 2018.

\bibitem{tang2016large}
Yuxing Tang, Josiah Wang, Boyang Gao, Emmanuel Dellandr{\'e}a, Robert
  Gaizauskas, and Liming Chen.
\newblock Large scale semi-supervised object detection using visual and
  semantic knowledge transfer.
\newblock In {\em CVPR}, pages 2119--2128, 2016.

\bibitem{tran2015learning}
Du~Tran, Lubomir Bourdev, Rob Fergus, Lorenzo Torresani, and Manohar Paluri.
\newblock Learning spatiotemporal features with 3d convolutional networks.
\newblock In {\em ICCV}, pages 4489--4497, 2015.

\bibitem{tzeng2017adversarial}
Eric Tzeng, Judy Hoffman, Kate Saenko, and Trevor Darrell.
\newblock Adversarial discriminative domain adaptation.
\newblock In {\em CVPR}, pages 7167--7176, 2017.

\bibitem{wang2018spotting}
Huiyun Wang, Youjiang Xu, and Yahong Han.
\newblock Spotting and aggregating salient regions for video captioning.
\newblock In {\em ACM MM}, pages 1519--1526, 2018.

\bibitem{wang2018deepda}
Mei Wang and Weihong Deng.
\newblock Deep visual domain adaptation: A survey.
\newblock {\em Neurocomputing}, 312:135--153, 2018.

\bibitem{wang2018revisiting}
Wenguan Wang, Jianbing Shen, Fang Guo, Ming-Ming Cheng, and Ali Borji.
\newblock Revisiting video saliency: A large-scale benchmark and a new model.
\newblock In {\em CVPR}, pages 4894--4903, 2018.

\bibitem{wang2017video}
Wenguan Wang, Jianbing Shen, and Ling Shao.
\newblock Video salient object detection via fully convolutional networks.
\newblock {\em IEEE Transactions on Image Processing}, 27(1):38--49, 2017.

\bibitem{wang2019revisiting}
Wenguan Wang, Jianbing Shen, Jianwen Xie, Ming-Ming Cheng, Haibin Ling, and Ali
  Borji.
\newblock Revisiting video saliency prediction in the deep learning era.
\newblock {\em IEEE transactions on pattern analysis and machine intelligence},
  43(1):220--237, 2019.

\bibitem{wang2018non}
Xiaolong Wang, Ross Girshick, Abhinav Gupta, and Kaiming He.
\newblock Non-local neural networks.
\newblock In {\em CVPR}, pages 7794--7803, 2018.

\bibitem{wang2018deep}
Jianbing~Shen Wenguan~Wang.
\newblock Deep visual attention prediction.
\newblock {\em IEEE Transactions on Image Processing}, 2018.

\bibitem{wusalsac}
Xinyi Wu, Zhenyao Wu, Jinglin Zhang, Lili Ju, and Song Wang.
\newblock Salsac: A video saliency prediction model with shuffled attentions
  and correlation-based convlstm.
\newblock In {\em AAAI}, pages 12410--12417, 2020.

\bibitem{xie2018rethinking}
Saining Xie, Chen Sun, Jonathan Huang, Zhuowen Tu, and Kevin Murphy.
\newblock Rethinking spatiotemporal feature learning: Speed-accuracy trade-offs
  in video classification.
\newblock In {\em ECCV}, pages 305--321, 2018.

\bibitem{yubing2011spatiotemporal}
Tong Yubing, Faouzi~Alaya Cheikh, Fahad Fazal~Elahi Guraya, Hubert Konik, and
  Alain Tr{\'e}meau.
\newblock A spatiotemporal saliency model for video surveillance.
\newblock {\em Cognitive Computation}, 3(1):241--263, 2011.

\bibitem{zhang2018deep}
Jing Zhang, Tong Zhang, Yuchao Dai, Mehrtash Harandi, and Richard Hartley.
\newblock Deep unsupervised saliency detection: A multiple noisy labeling
  perspective.
\newblock In {\em CVPR}, pages 9029--9038, 2018.

\bibitem{zhang2017amulet}
Pingping Zhang, Dong Wang, Huchuan Lu, Hongyu Wang, and Xiang Ruan.
\newblock Amulet: Aggregating multi-level convolutional features for salient
  object detection.
\newblock In {\em {IEEE} International Conference on Computer Vision}, 2017.

\bibitem{zhang2017curriculum}
Yang Zhang, Philip David, and Boqing Gong.
\newblock Curriculum domain adaptation for semantic segmentation of urban
  scenes.
\newblock In {\em ICCV}, pages 2020--2030, 2017.

\end{thebibliography}
\end{document}